%% file: main.tex
\documentclass[acmtog,nonacm]{acmart}
\definecolor{ppoaverage}{HTML}{FFA300}
\definecolor{ppominimum}{HTML}{16834F}
\begin{document}

\title{Synthesizing Reactive Character Behaviors for Continuous Games via Programmatic Policy Search}

\author{Maxim Gumin}
\orcid{0009-0002-1204-6797}
\affiliation{\institution{Brown University}\city{Providence}\country{USA}}
\email{maxgumin@gmail.com}

\author{Hsueh-Ti Derek Liu}
\orcid{0009-0001-1753-4485}
\affiliation{\institution{Roblox}\city{Vancouver}\country{Canada}}
\email{hsuehtil@gmail.com}

\author{Victor Zordan}
\orcid{0000-0002-7309-7013}
\affiliation{\institution{Clemson University}\city{Clemson}\country{USA}}
\email{vbz@clemson.edu}

\author{Daniel Ritchie}
\orcid{0000-0002-8253-0069}
\affiliation{\institution{Brown University}\city{Providence}\country{USA}}
\email{daniel_ritchie@brown.edu}
\renewcommand{\shortauthors}{Gumin et al.}

\begin{abstract}
We present a method for synthesizing reactive character behaviors for continuous games as compact, human-readable programs. Game AI practice still relies heavily on manually authored behavior trees, state machines, and scripts, while academic reinforcement learning typically produces opaque neural controllers that are expensive to train and difficult to edit. Our approach bridges this gap by searching directly over a domain-specific language for continuous-space game policies. The language is designed around reactive geometric decisions and includes higher-order constructs such as direction maximization. These constructs help discretize a continuous behavior space into enumerable program structures. To make program search practical, we introduce a large set of synthesis antipatterns that remove redundant program forms while preserving behavioral coverage. We further combine bottom-up symbolic enumeration with top-down guidance from a coding agent. Our resulting method, \emph{agentic sketching}, has the agent propose high-level policy structure and call an enumerator to complete local program slots. We evaluate the method on a benchmark of 14 continuous games, ranging from classic control tasks to multi-agent football. We find that pure enumeration is often more efficient than using a coding agent alone, while the combined method substantially outperforms both. Our results suggest that programmatic policy search can be a practical authoring tool for game AI: designers specify reward functions, and the system discovers editable behaviors that are effective, portable, and often surprising.
\end{abstract}



\begin{teaserfigure}
  \centering
  \includegraphics[width=\linewidth]{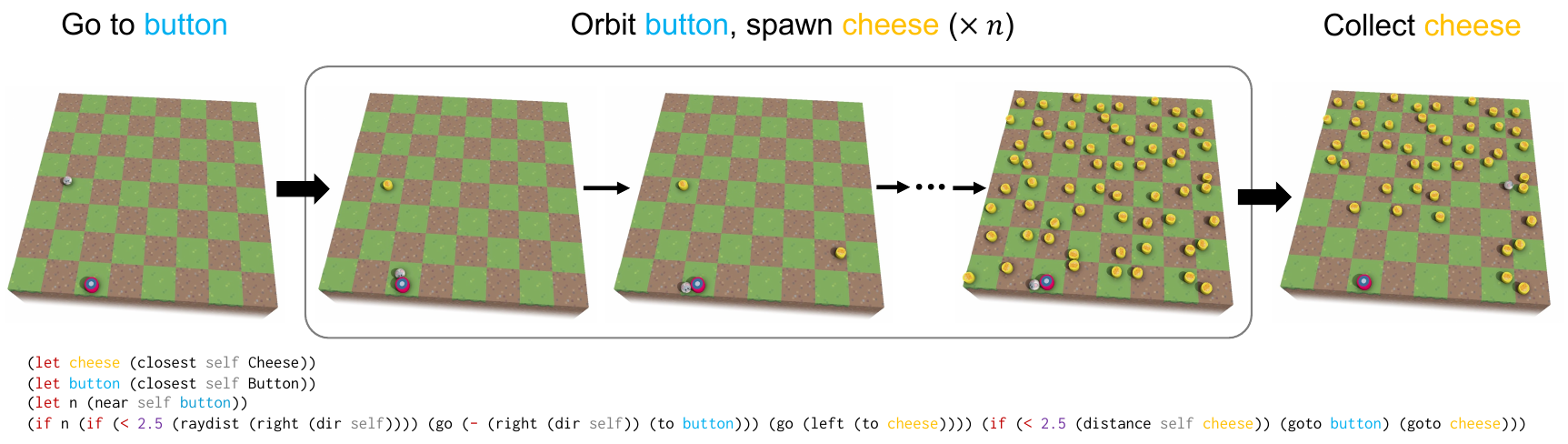}
  \caption{\emph{(Top)} The behavior of a mouse character in our ButtonCheese game, using a program synthesized by our agentic sketching method. The synthesizer discovers a surprising and effective policy in which the mouse goes to the button, presses it repeatedly to fill the map with cheese, and then rapidly collects the cheese. \emph{(Bottom)} The last line is the automatically synthesized policy in our character behavior domain-specific language; the first three lines define game-provided sensor functions.}
  \Description{A ButtonCheese game sequence shows the mouse repeatedly pressing a button to spawn many cheese objects and then collecting them, followed by the synthesized policy program.}
  \label{fig:teaser}
\end{teaserfigure}

\maketitle

\input{1-Introduction}
\input{2-RelatedWork}
\input{3-Benchmark}
\input{4-DSL}
\input{5-Algorithm}
\input{6-EvaluationAndResults}
\input{7-Conclusion}

\clearpage

\begin{figure*}[p]
  \centering
  \includegraphics[width=0.32\textwidth]{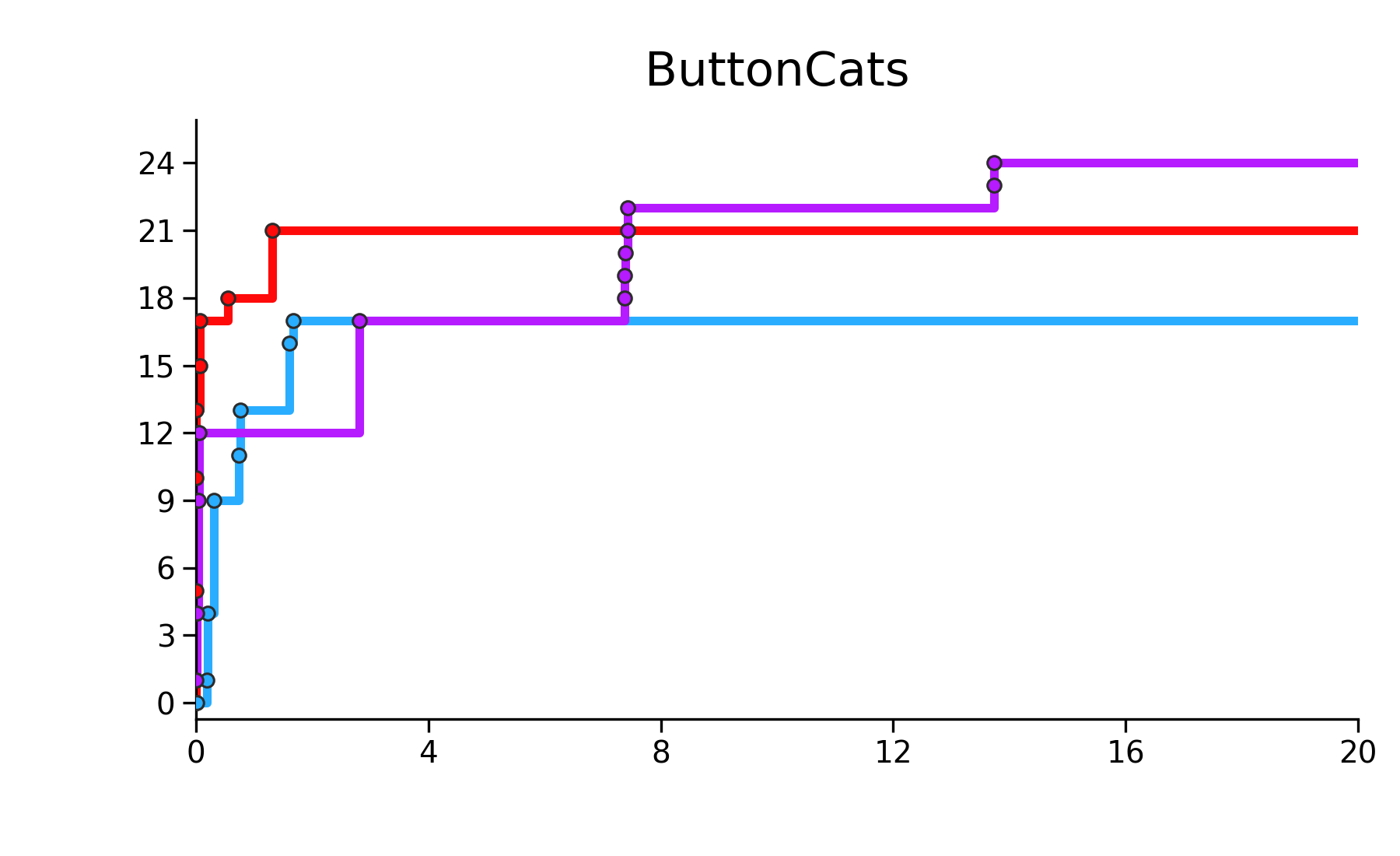}\hfill
  \includegraphics[width=0.32\textwidth]{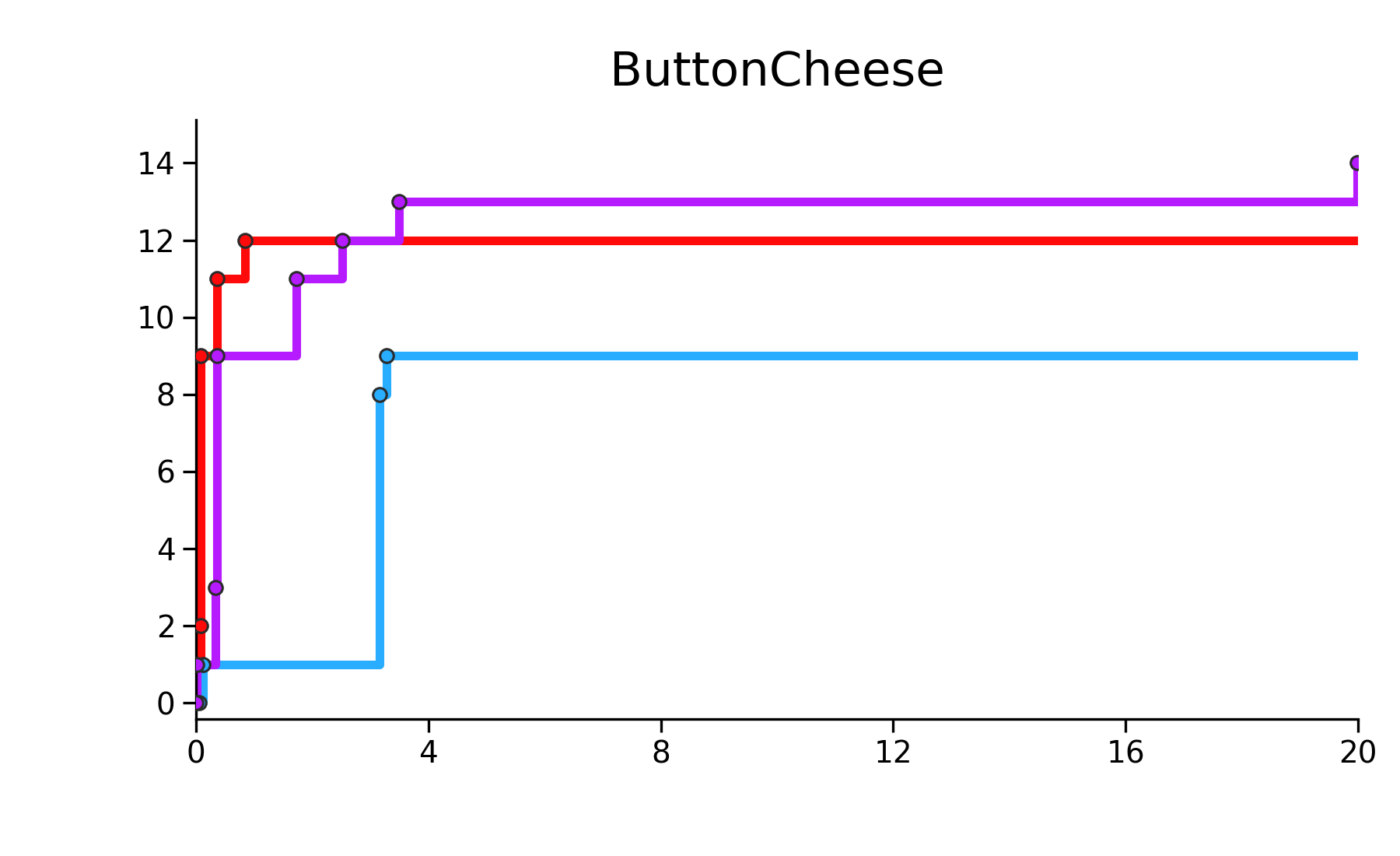}\hfill
  \includegraphics[width=0.32\textwidth]{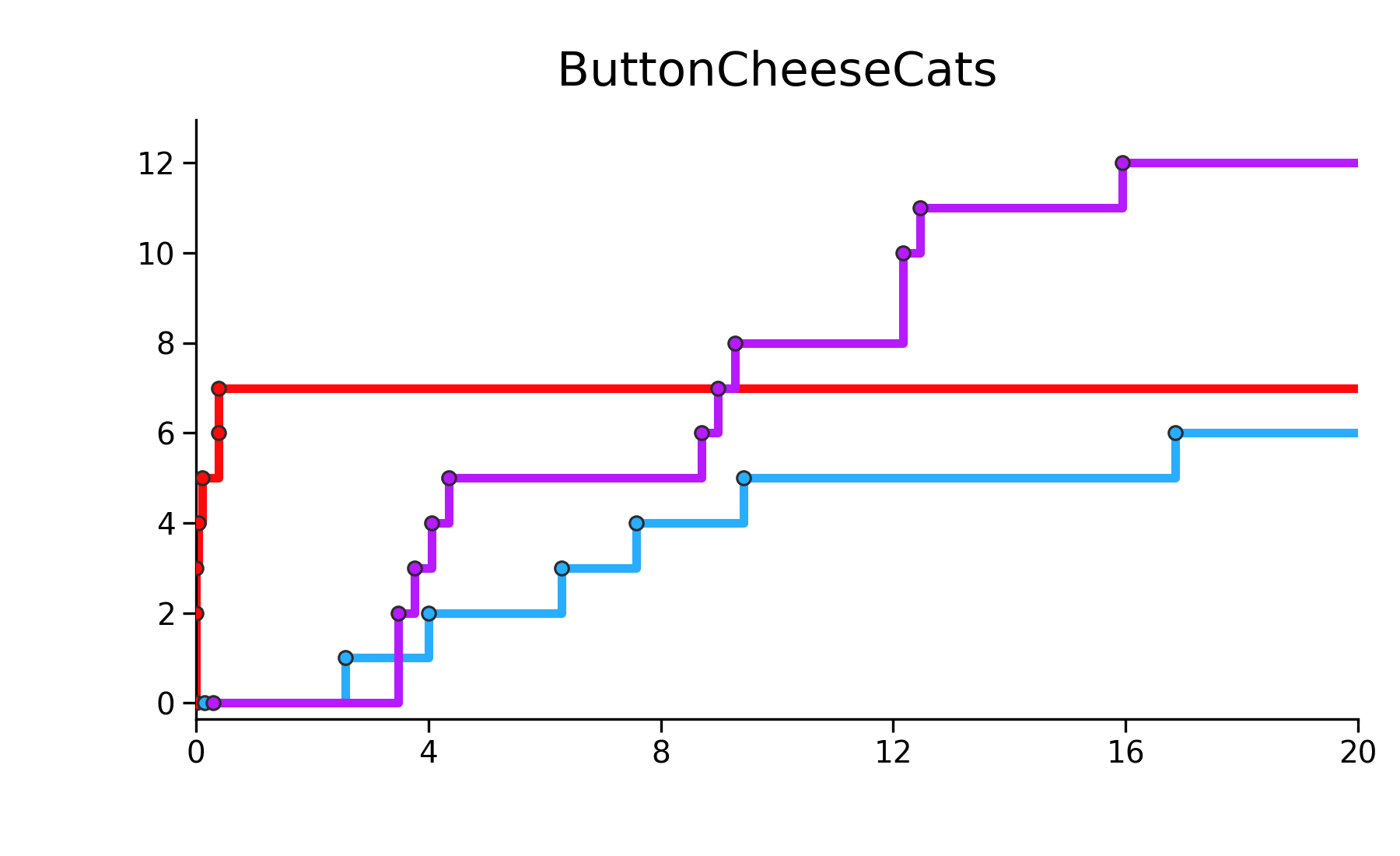}

  \includegraphics[width=0.32\textwidth]{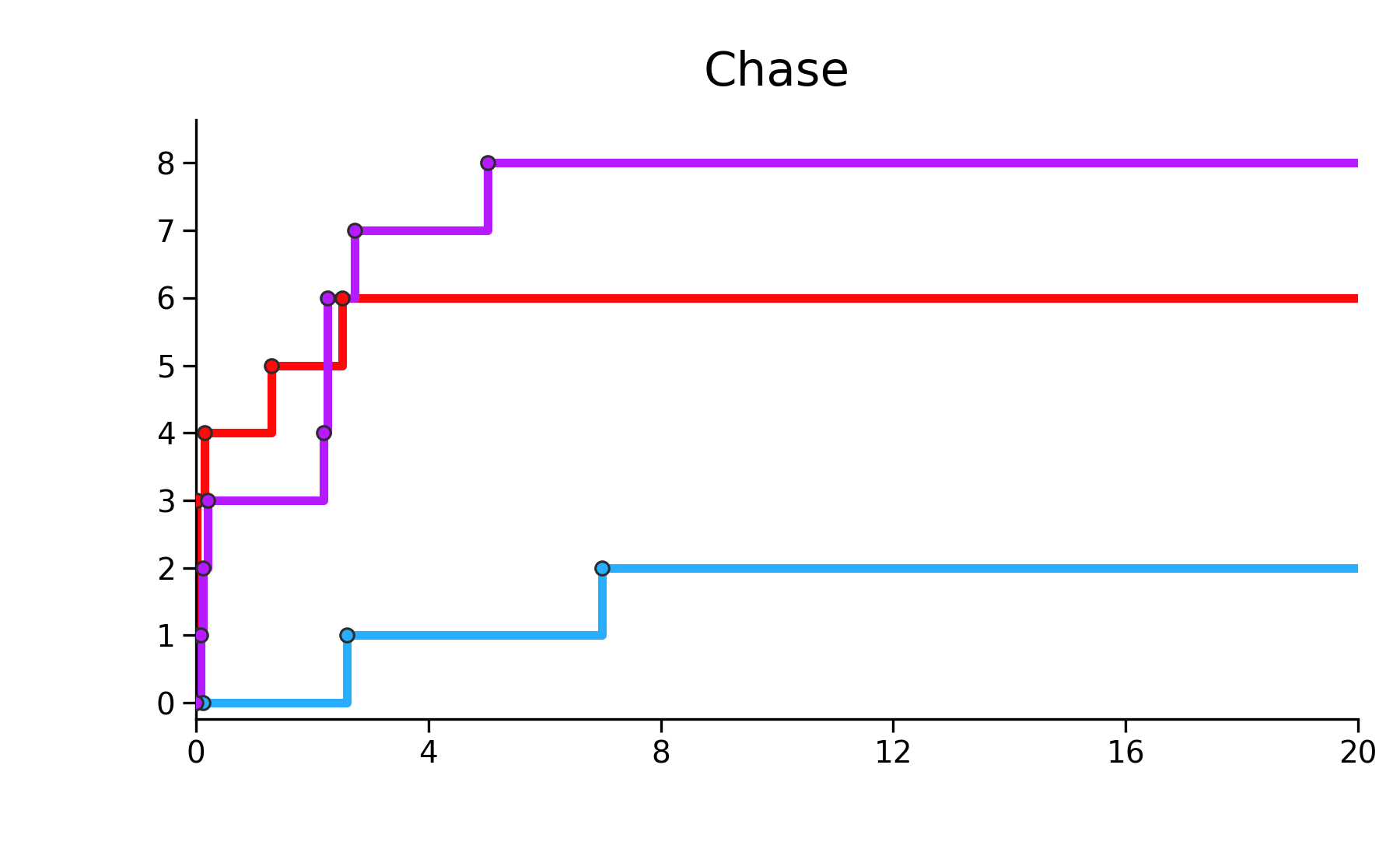}\hfill
  \includegraphics[width=0.32\textwidth]{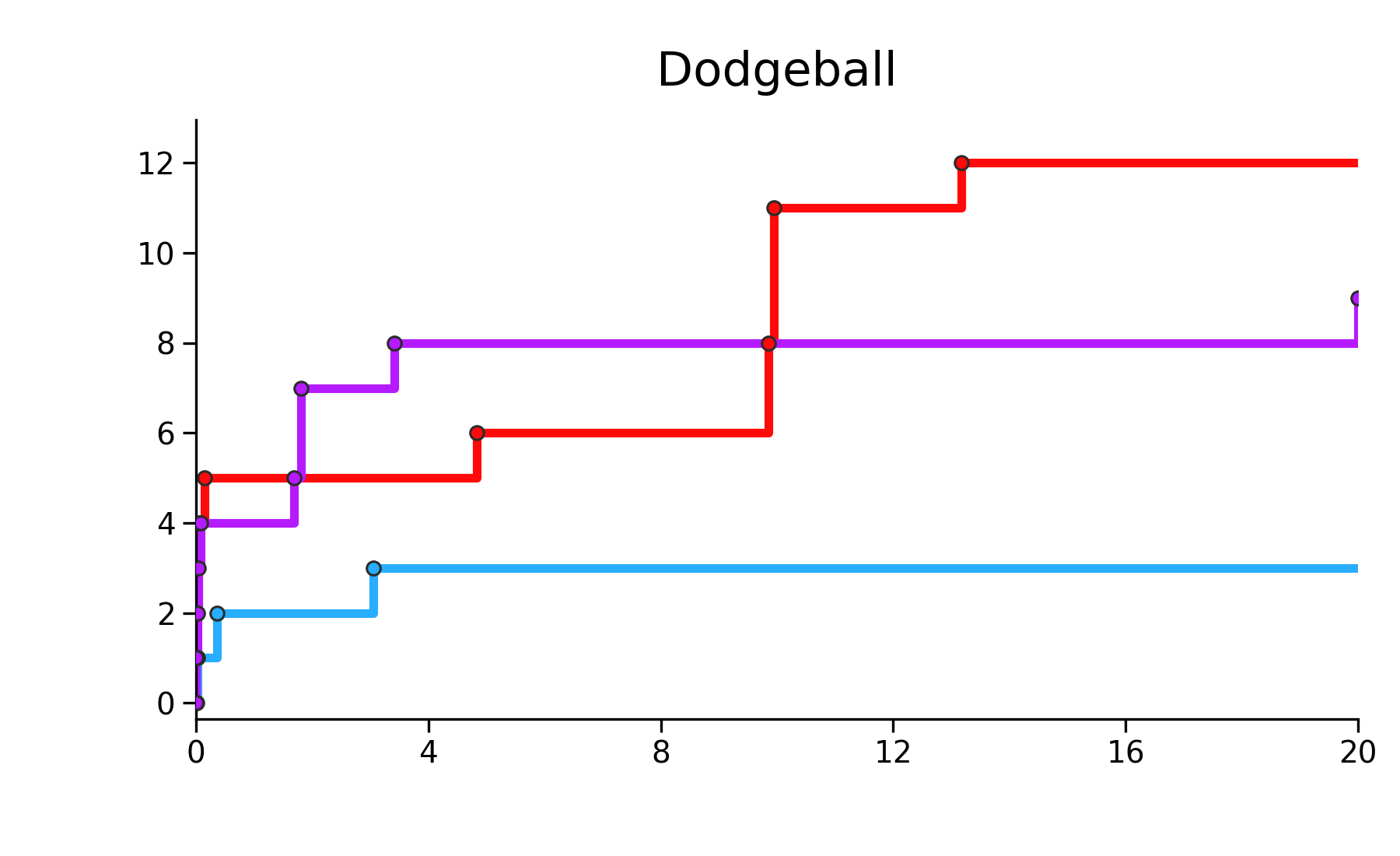}\hfill
  \includegraphics[width=0.32\textwidth]{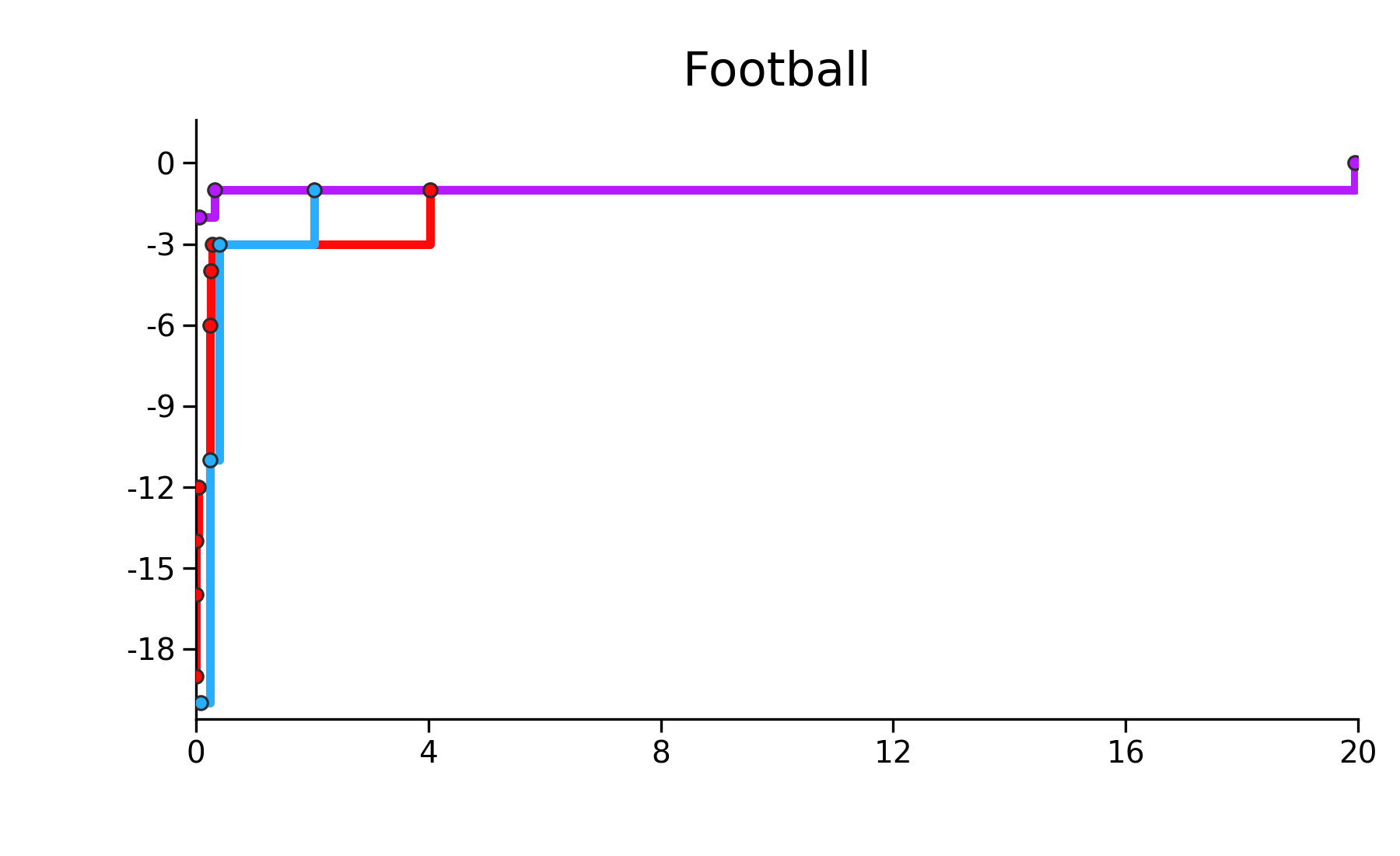}

  \includegraphics[width=0.32\textwidth]{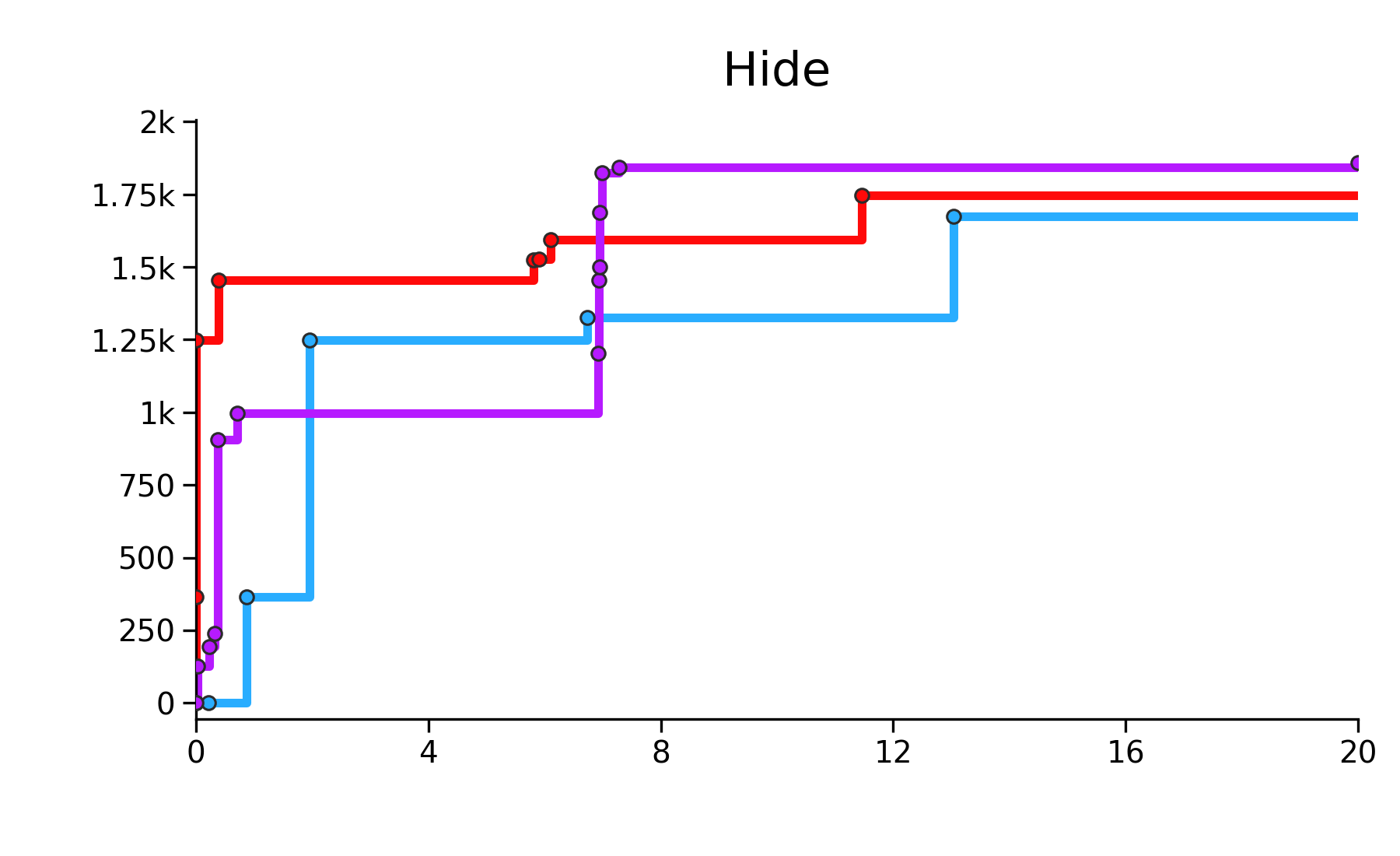}\hfill
  \includegraphics[width=0.32\textwidth]{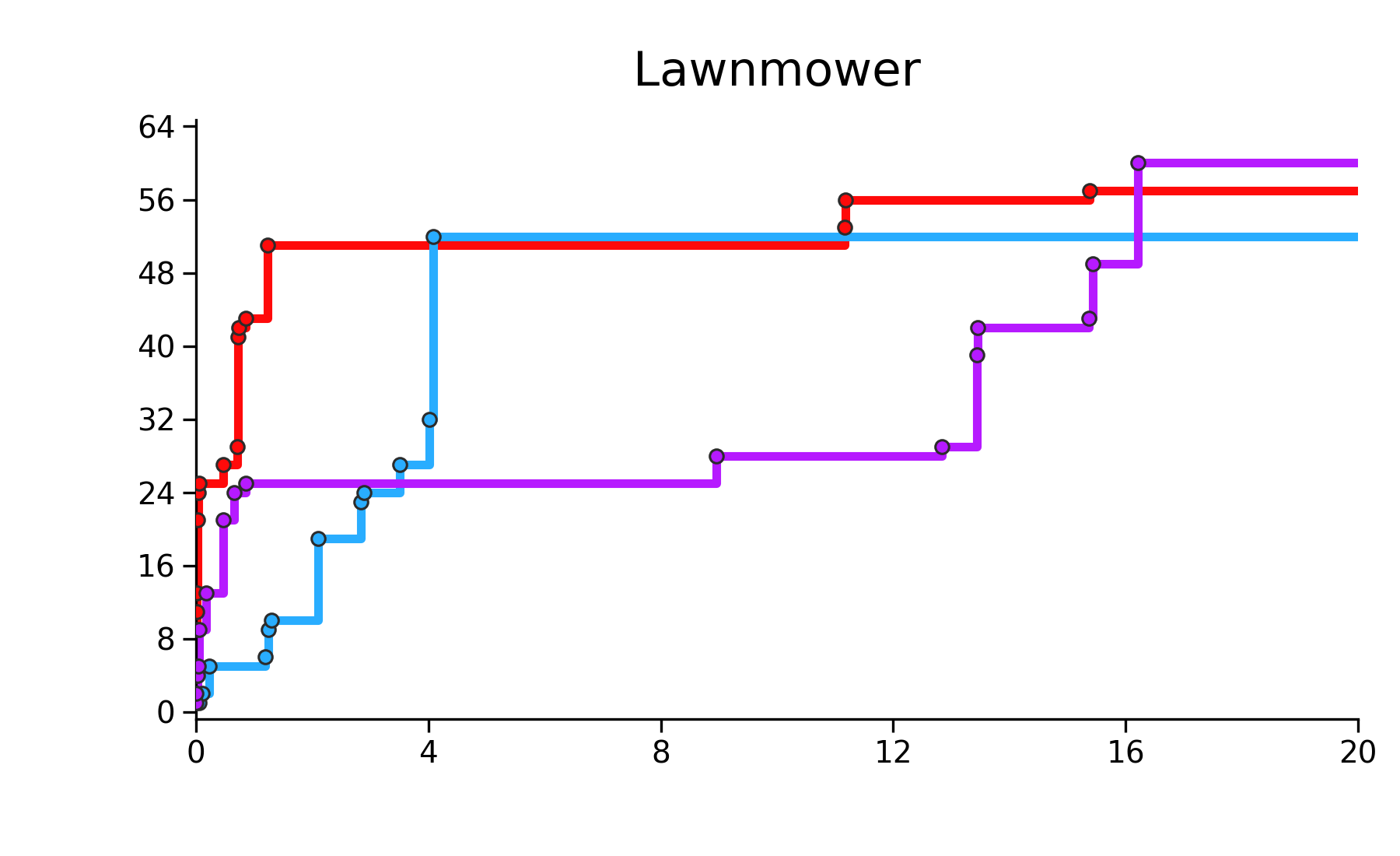}\hfill
  \includegraphics[width=0.32\textwidth]{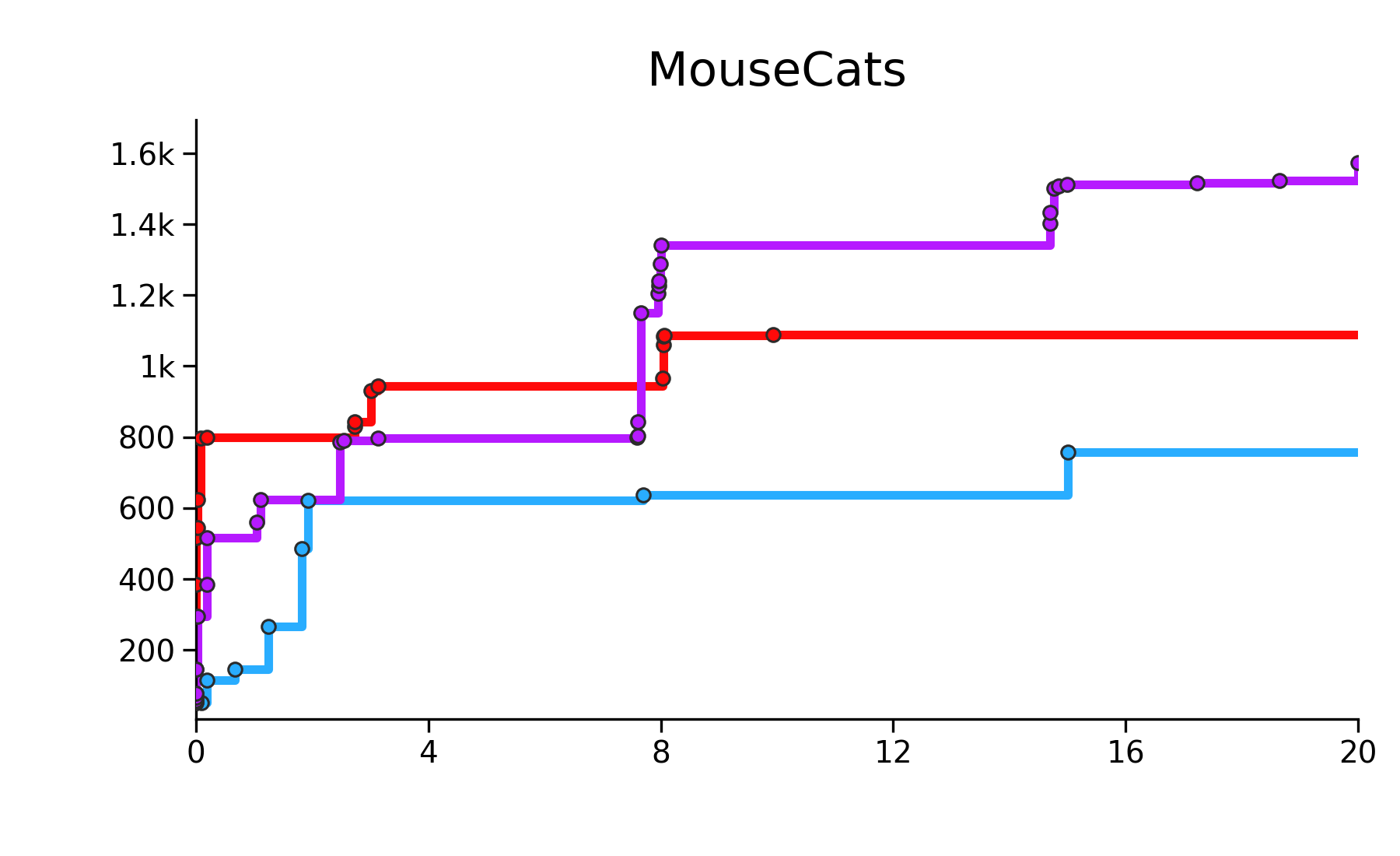}

  \includegraphics[width=0.32\textwidth]{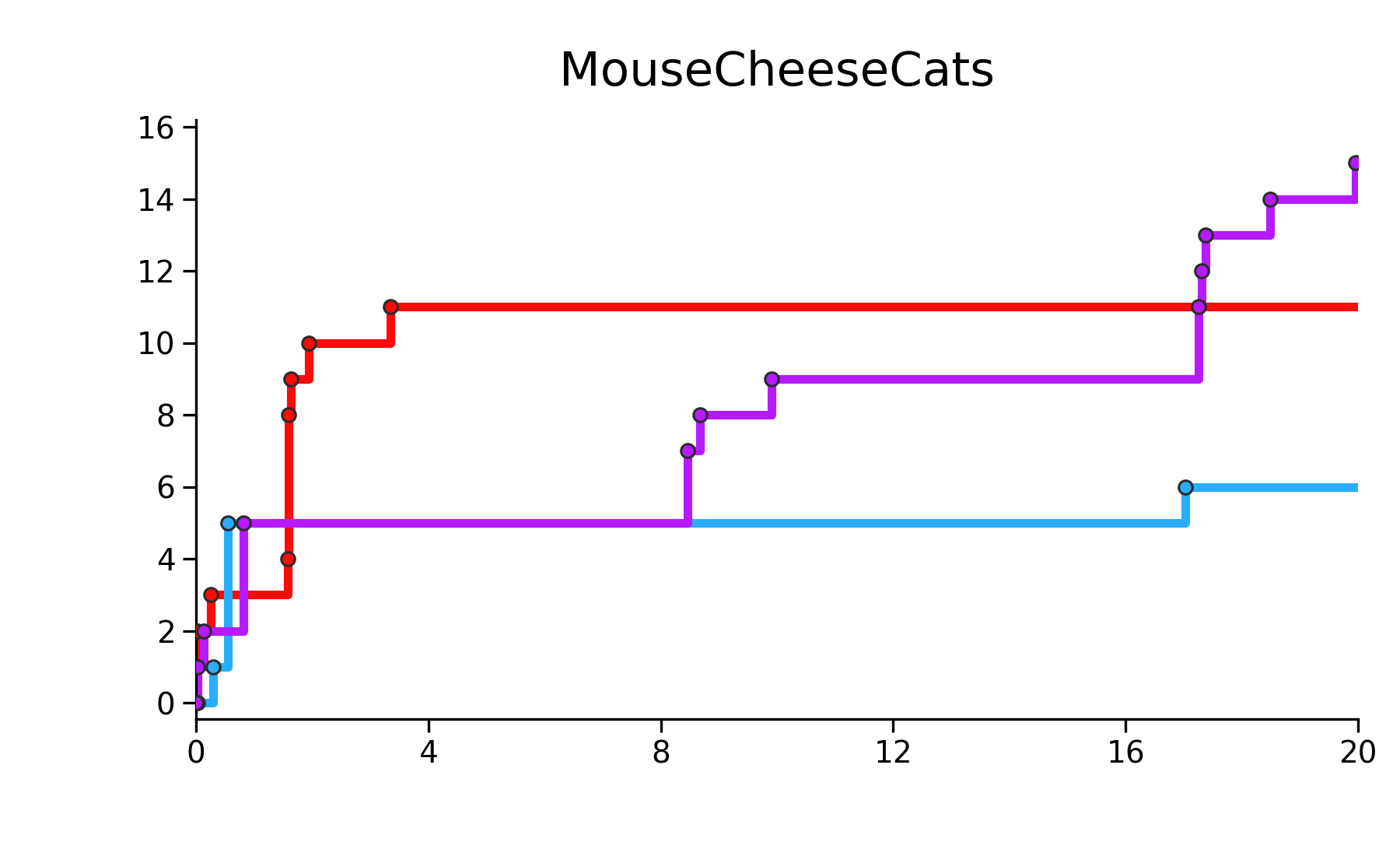}\hfill
  \includegraphics[width=0.32\textwidth]{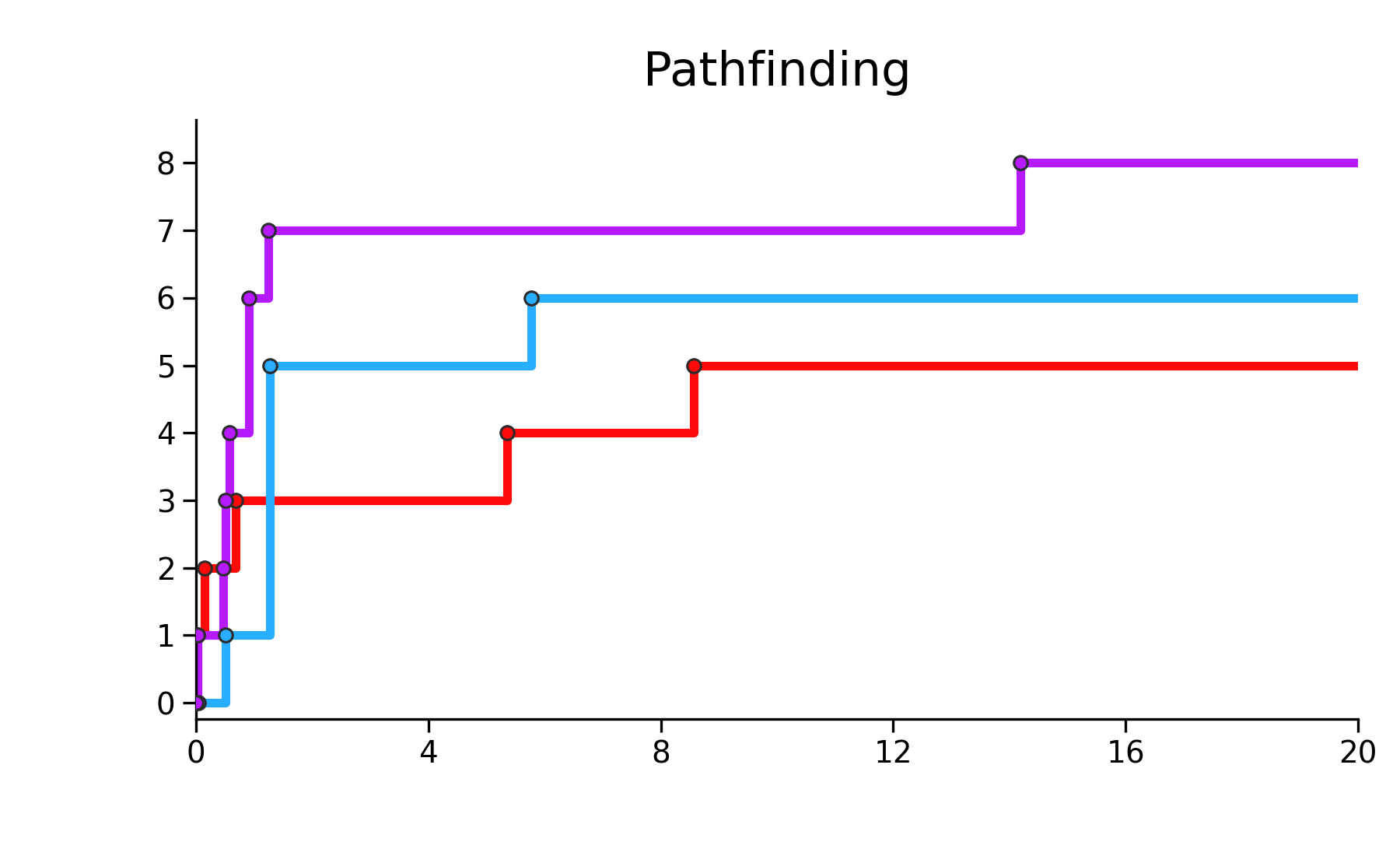}\hfill
  \includegraphics[width=0.32\textwidth]{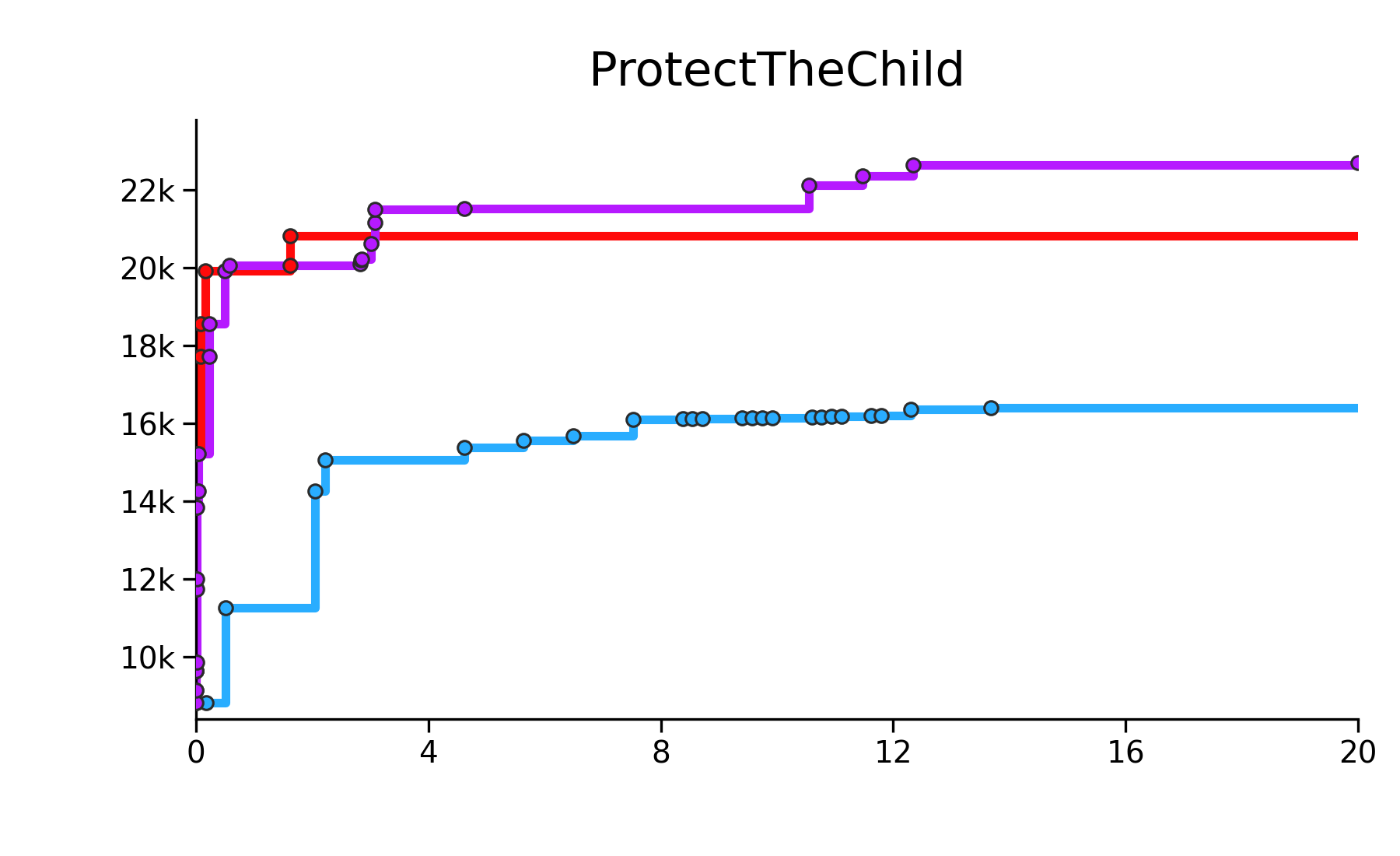}

  \includegraphics[width=0.32\textwidth]{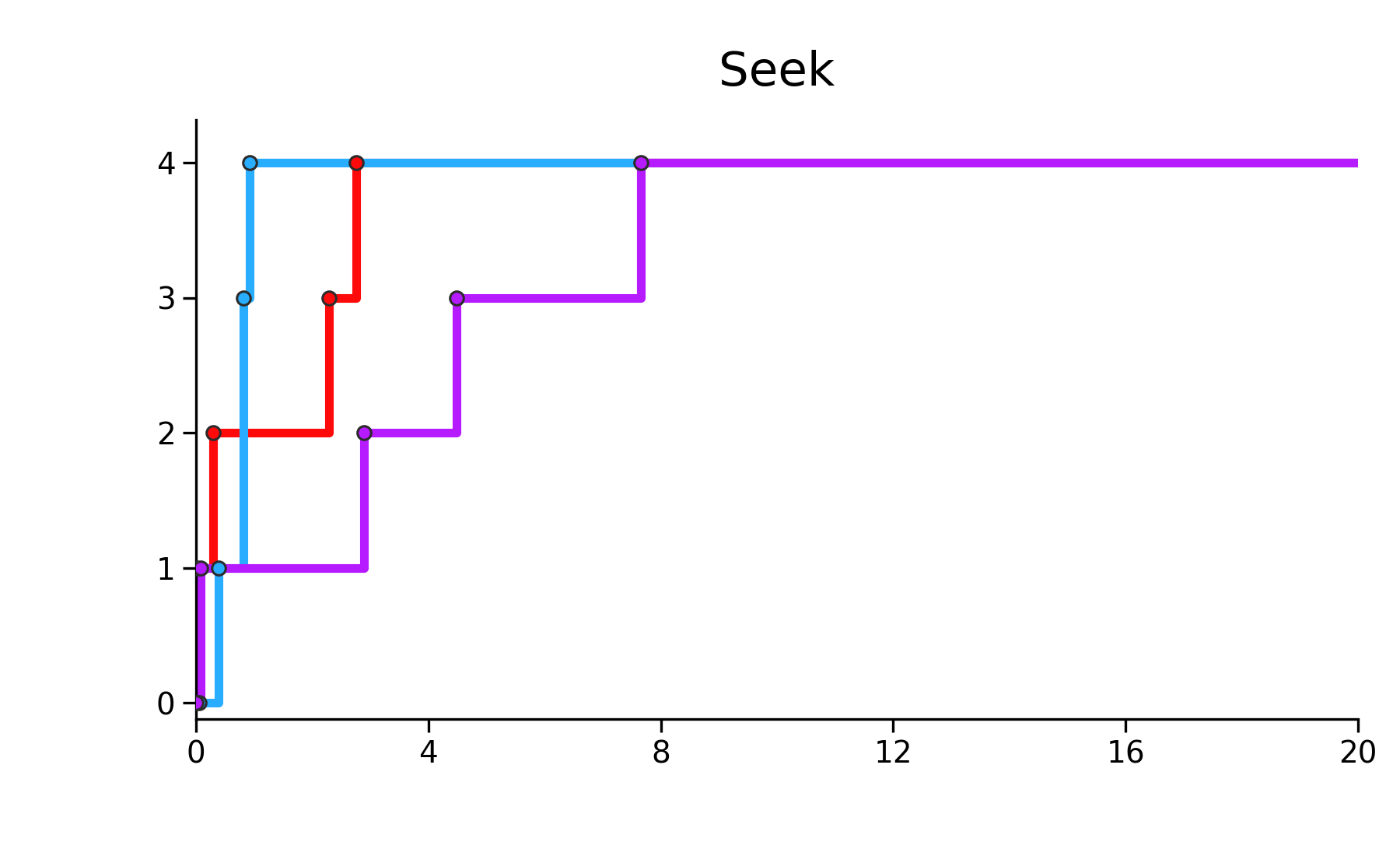}\hfill
  \includegraphics[width=0.32\textwidth]{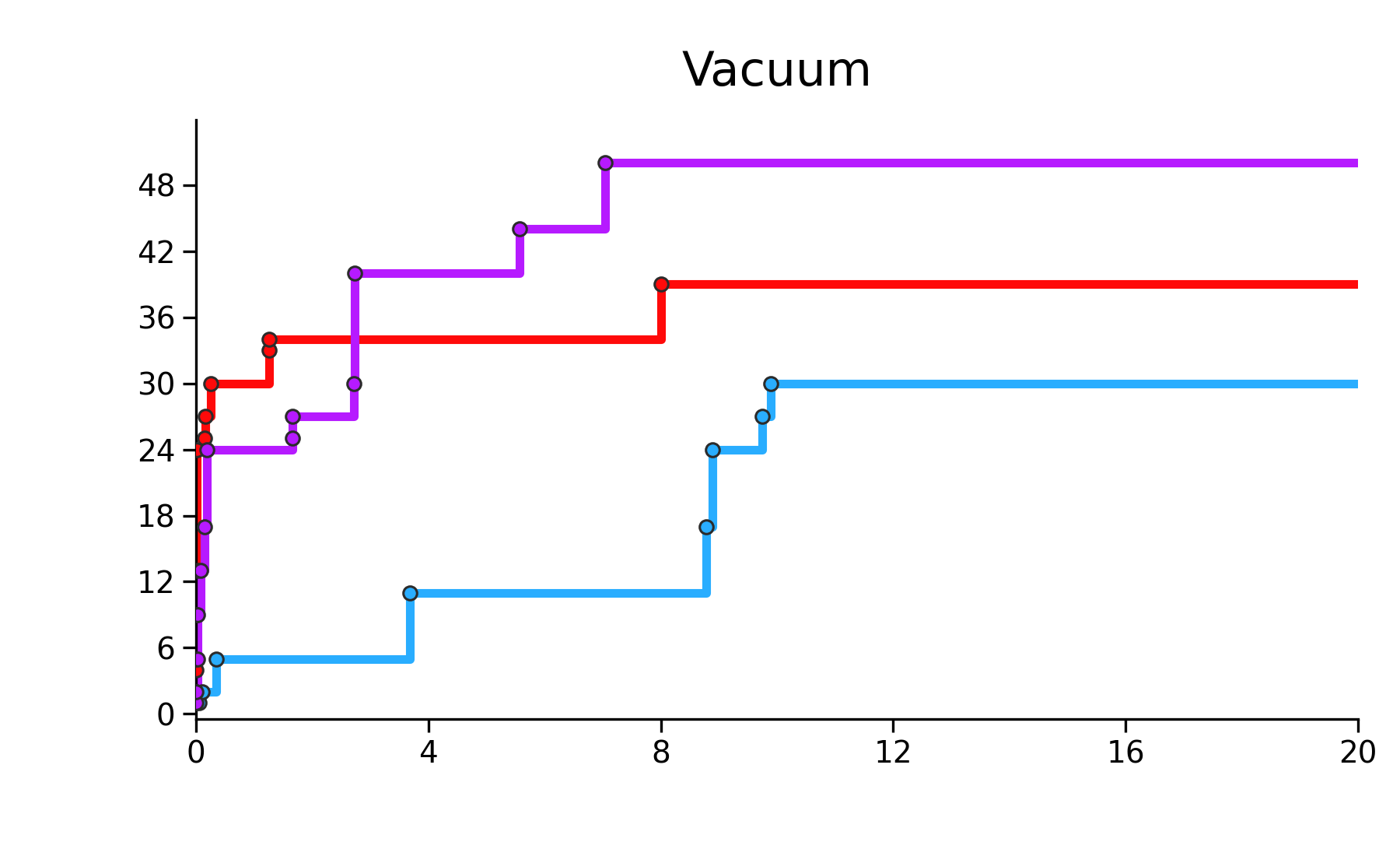}\hfill
  \makebox[0.32\textwidth]{}
  \caption{Search progress over 20 minutes for all benchmark games. {\color{red}{Red}} shows pure bottom-up symbolic search, {\color{cyan}{cyan}} shows pure coding-agent search, and {\color{violet}{violet}} shows agentic sketching. In most games, the coding agent is eventually dominated by search, which is itself eventually dominated by agentic sketching.}
  \Description{Fourteen plots compare the best score found over 20 minutes by bottom-up symbolic search, coding-agent search, and agentic sketching.}
  \label{fig:search-history}
\end{figure*}
\clearpage

\begin{figure*}[p]
  \centering
  \includegraphics[width=\textwidth,height=0.87\textheight,keepaspectratio]{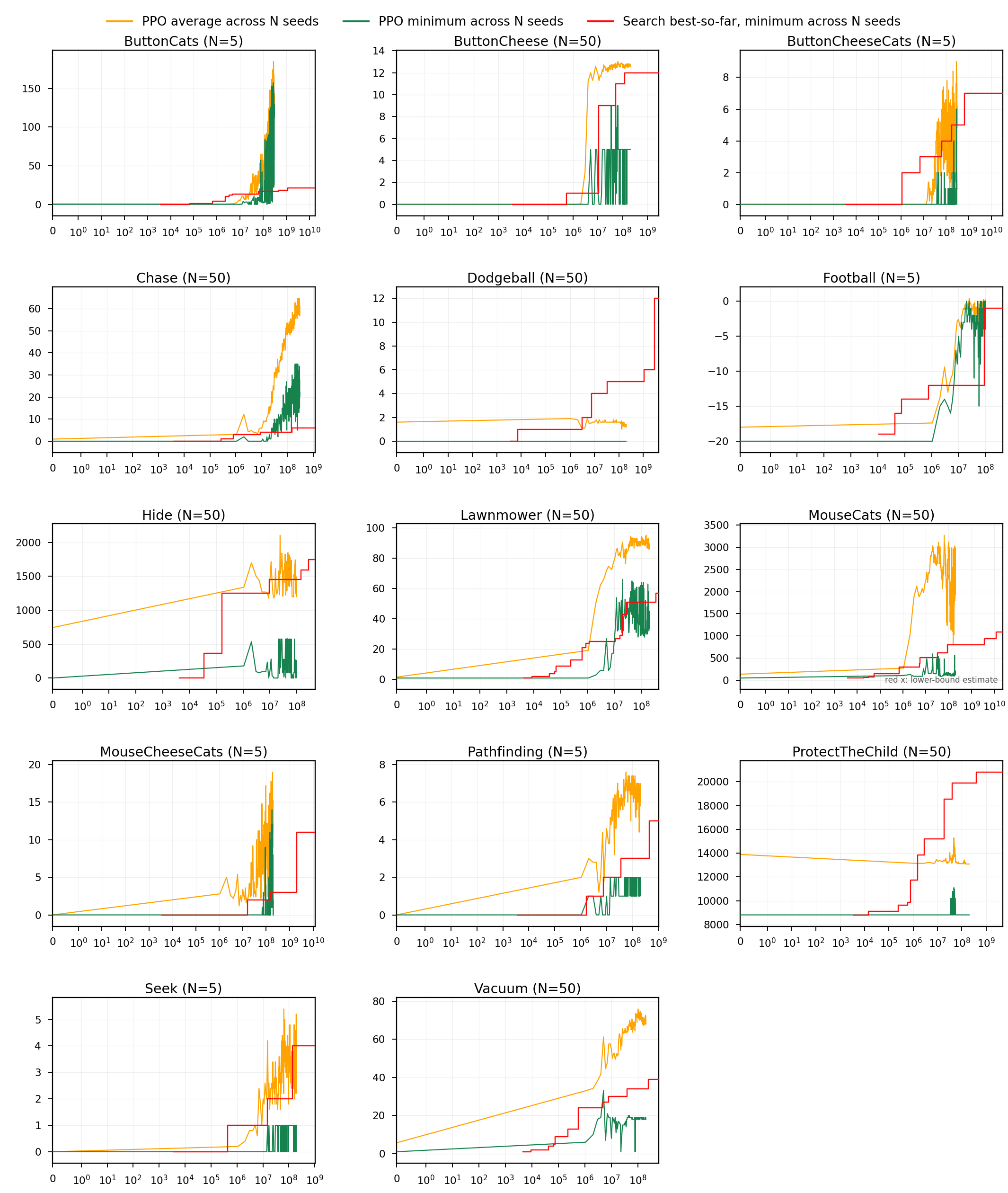}
  \caption{Fresh-seed PPO learning curves and bottom-up search progress for all benchmark games. \textcolor{ppoaverage}{Yellow} and \textcolor{ppominimum}{green} show the average and minimum test scores, respectively, of the current PPO checkpoint; \textcolor{red}{red} shows the minimum score of the best policy found so far by bottom-up search. The horizontal axis shows simulator frames on a symmetric-log scale.}
  \Description{Fourteen plots compare PPO average and minimum test scores with the minimum score found by bottom-up search as simulator frames increase.}
  \label{fig:ppo-search-history}
\end{figure*}
\clearpage
\begin{figure*}[p]
    \centering
    
    \setlength{\tabcolsep}{1pt}
    \begin{tabular}{@{}cccc@{}}
         \includegraphics[width=0.24\linewidth]{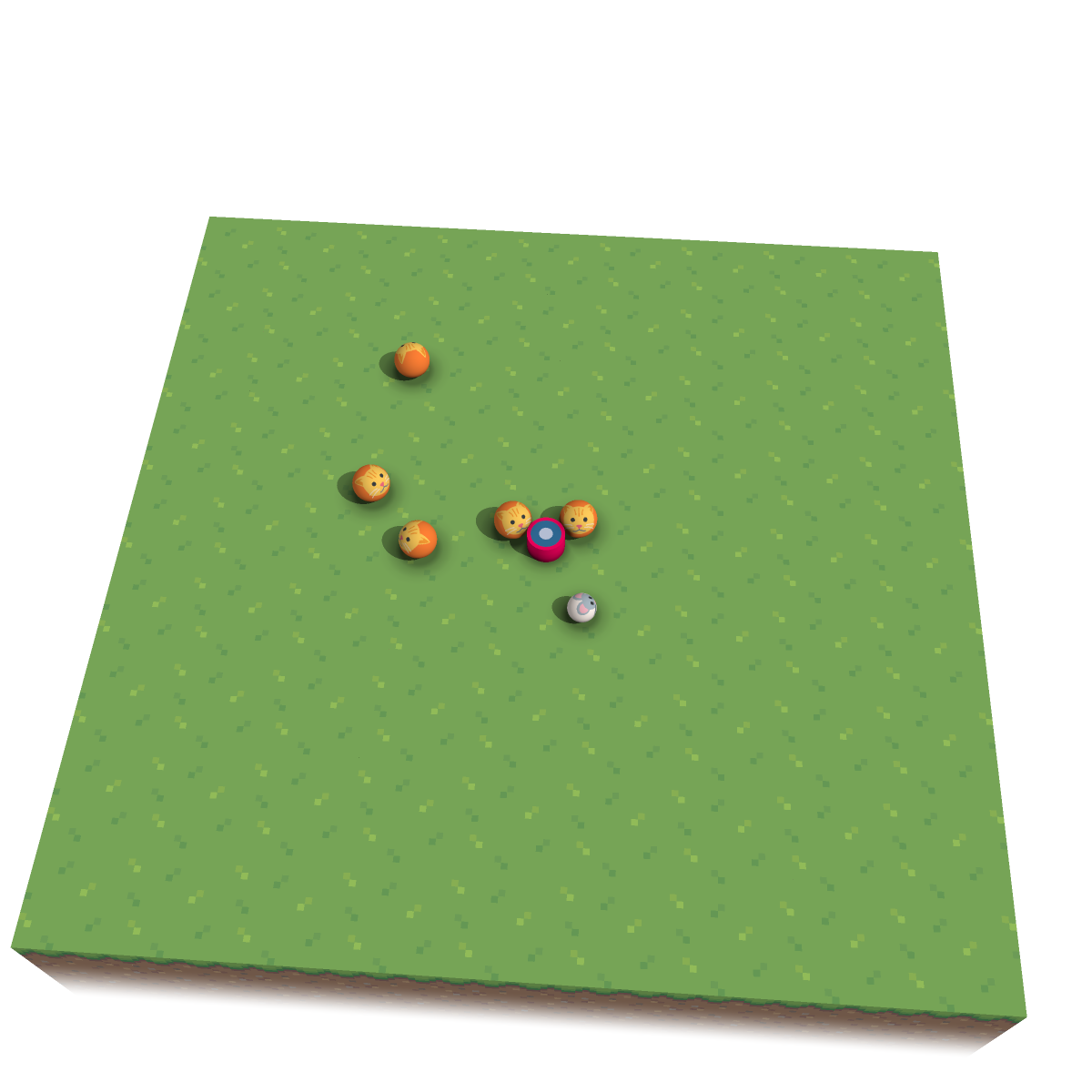}
         &
         \includegraphics[width=0.24\linewidth]{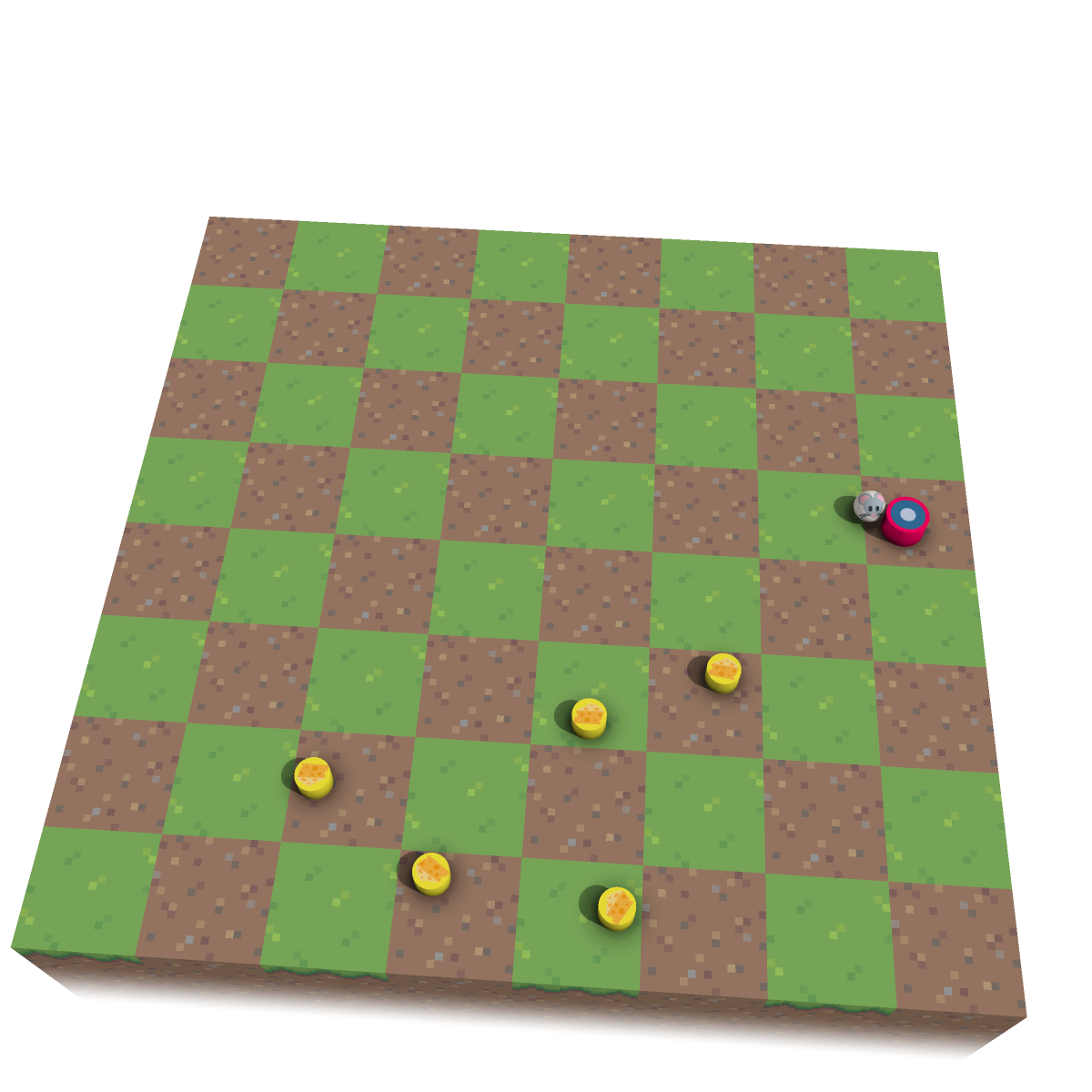}
         &
         \includegraphics[width=0.24\linewidth]{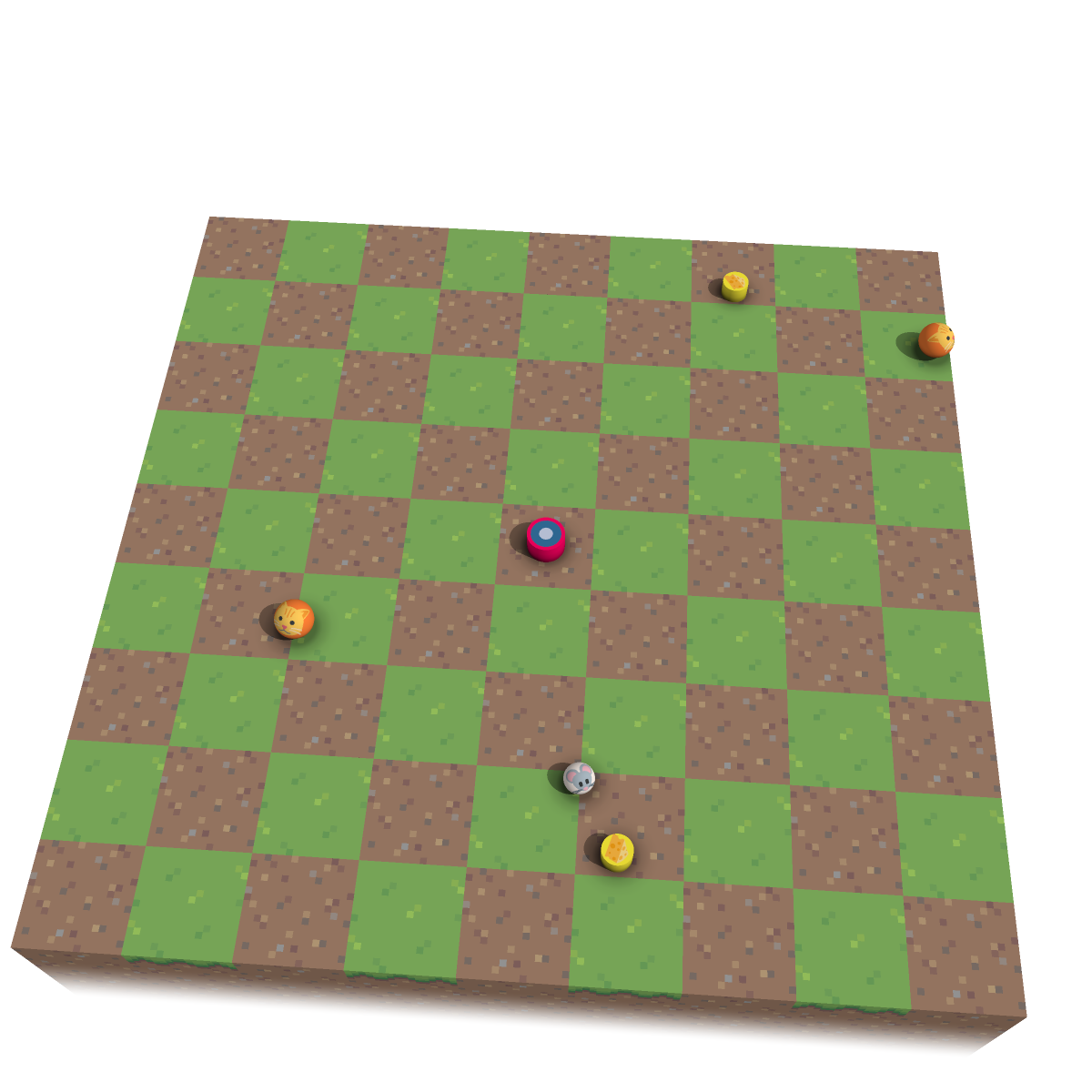}
         &
         \includegraphics[width=0.24\linewidth]{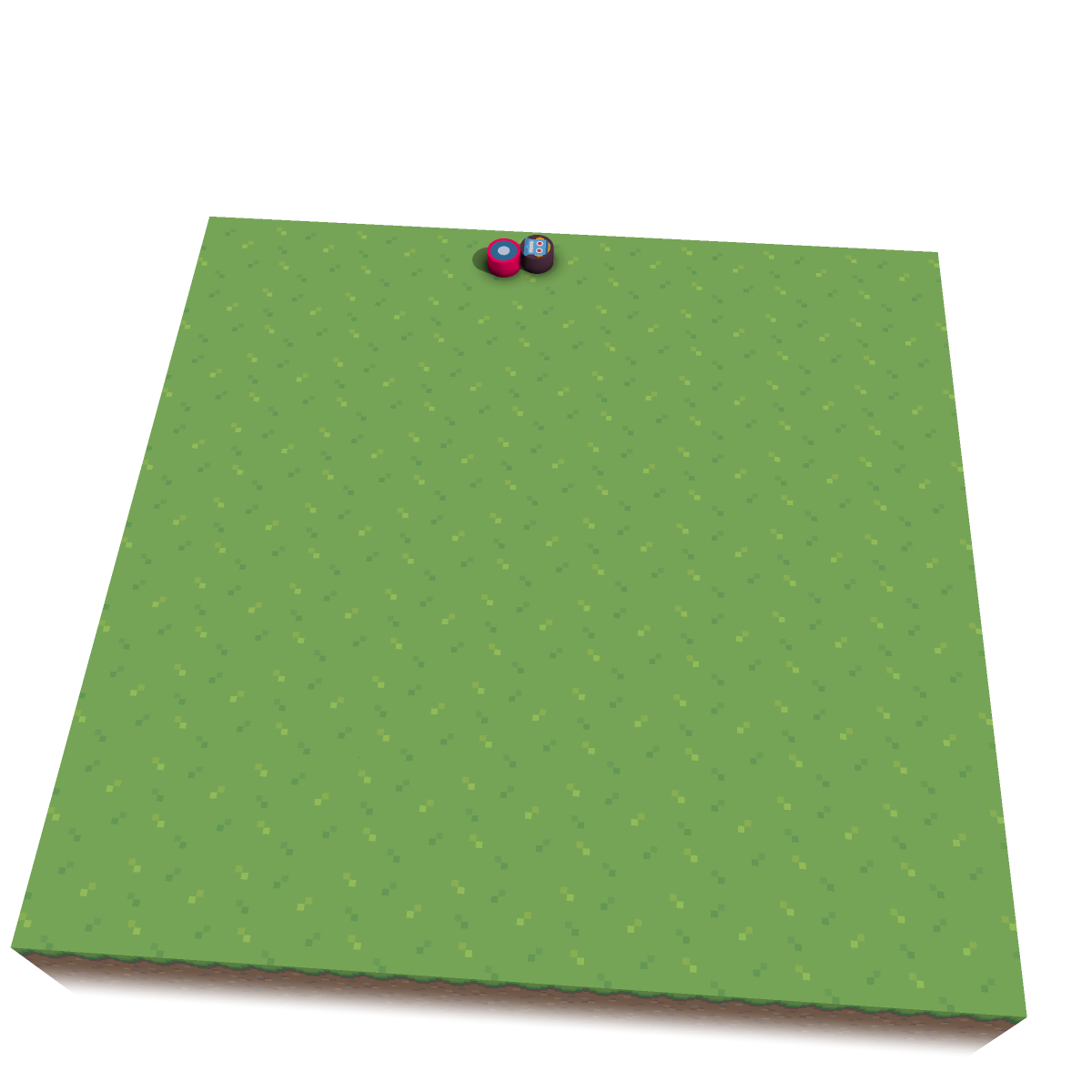}
         \\
         ButtonCats & ButtonCheese & ButtonCheeseCats & Chase
         \\
         \includegraphics[width=0.24\linewidth]{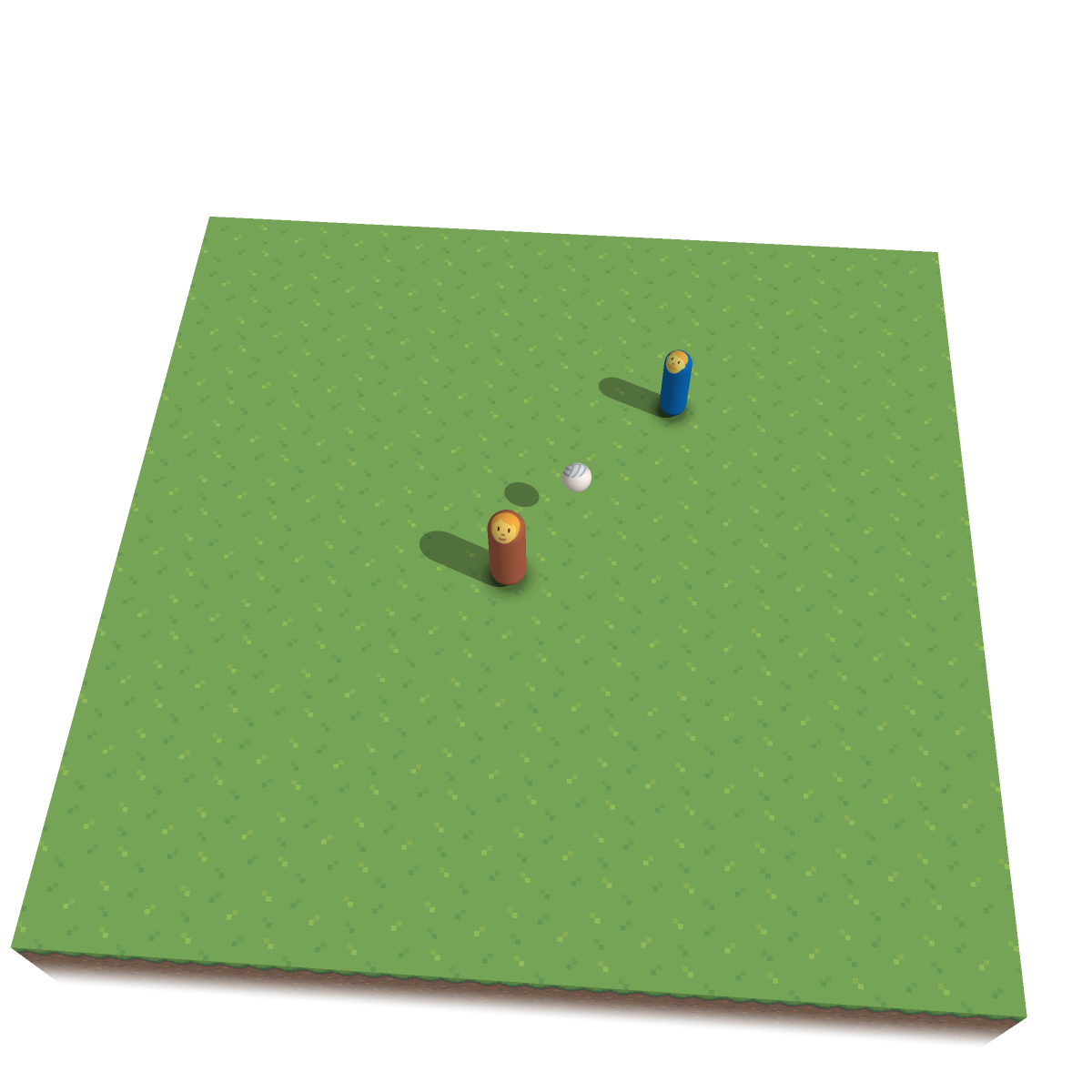}
         &
         \includegraphics[width=0.24\linewidth]{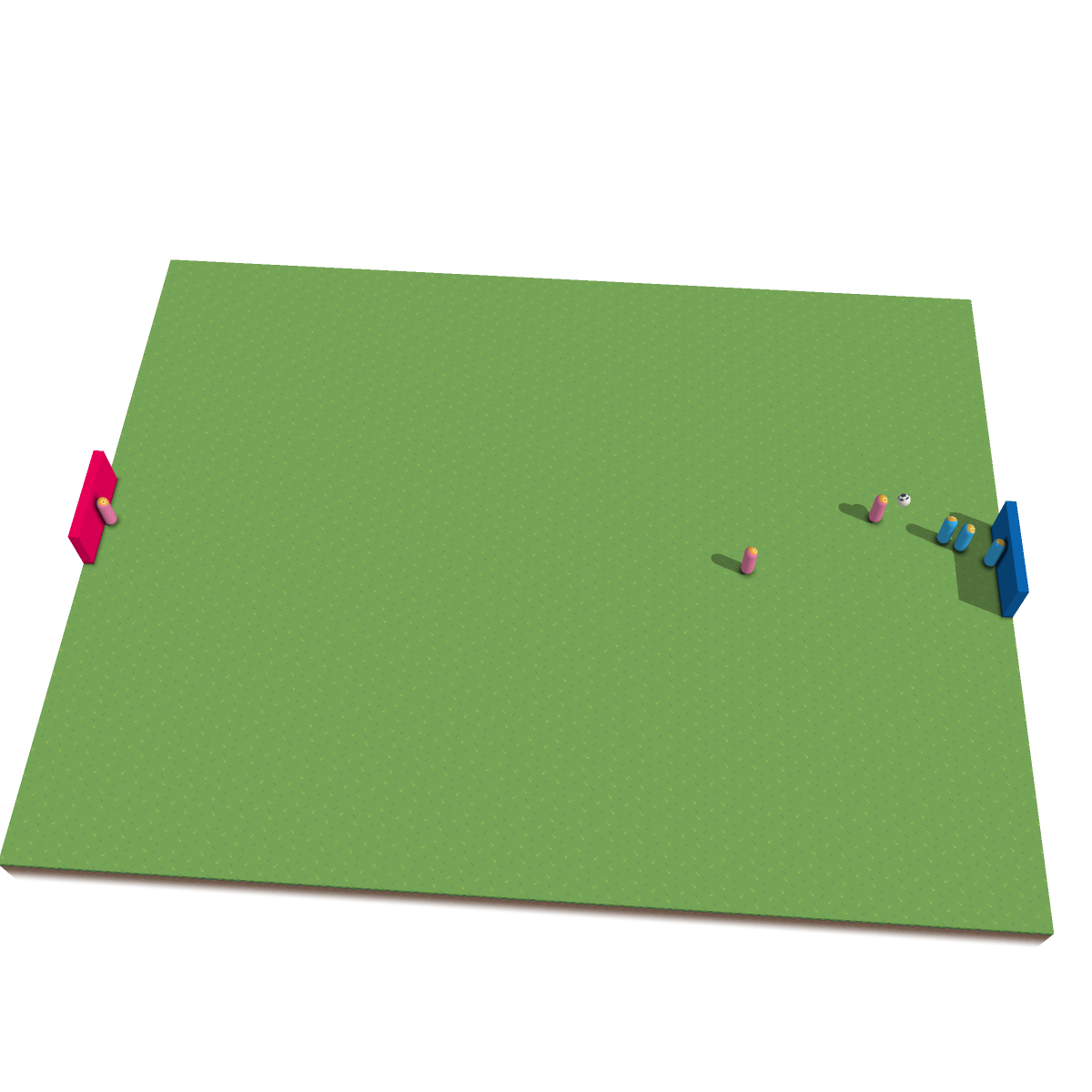}
         &
         \includegraphics[width=0.24\linewidth]{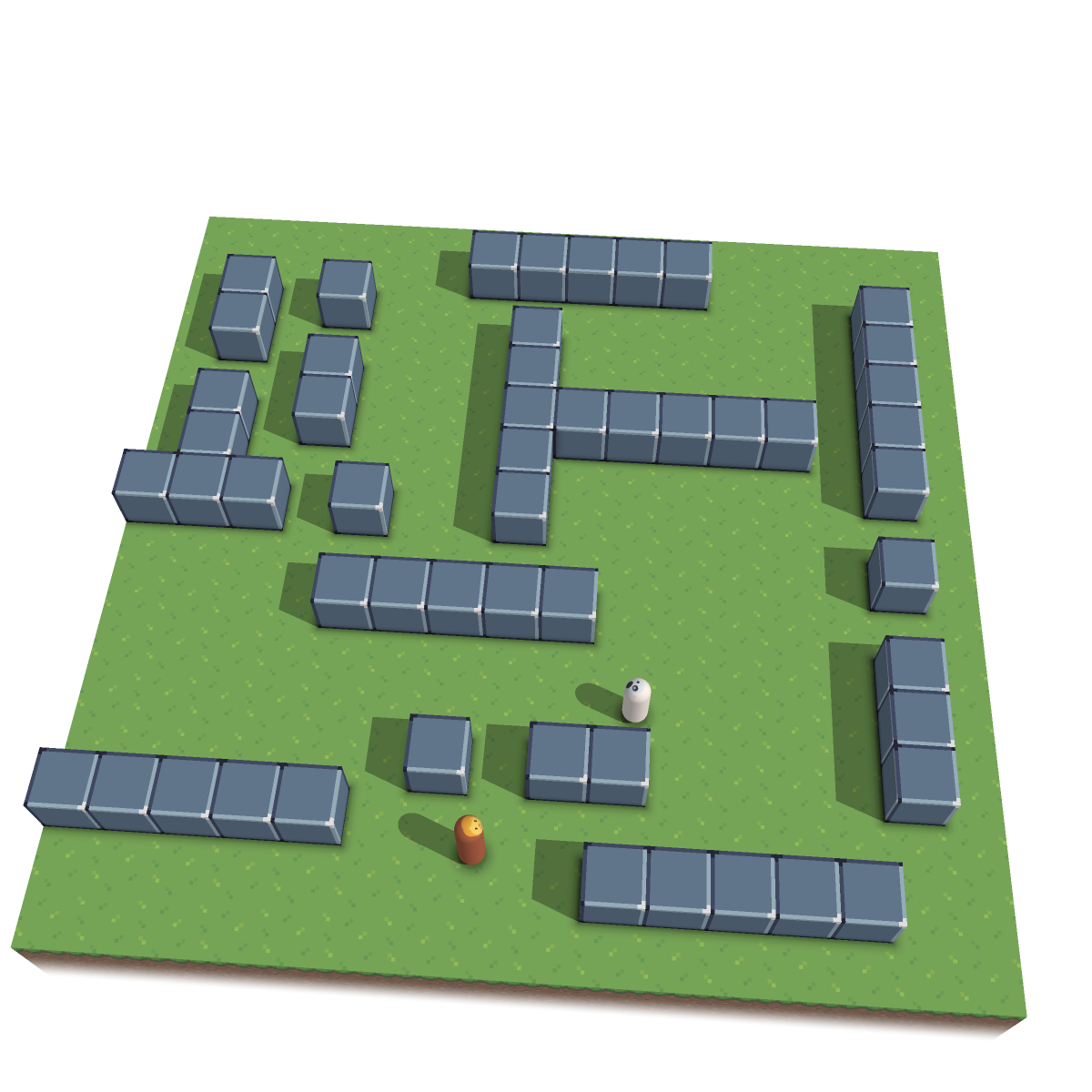}
         &
         \includegraphics[width=0.24\linewidth]{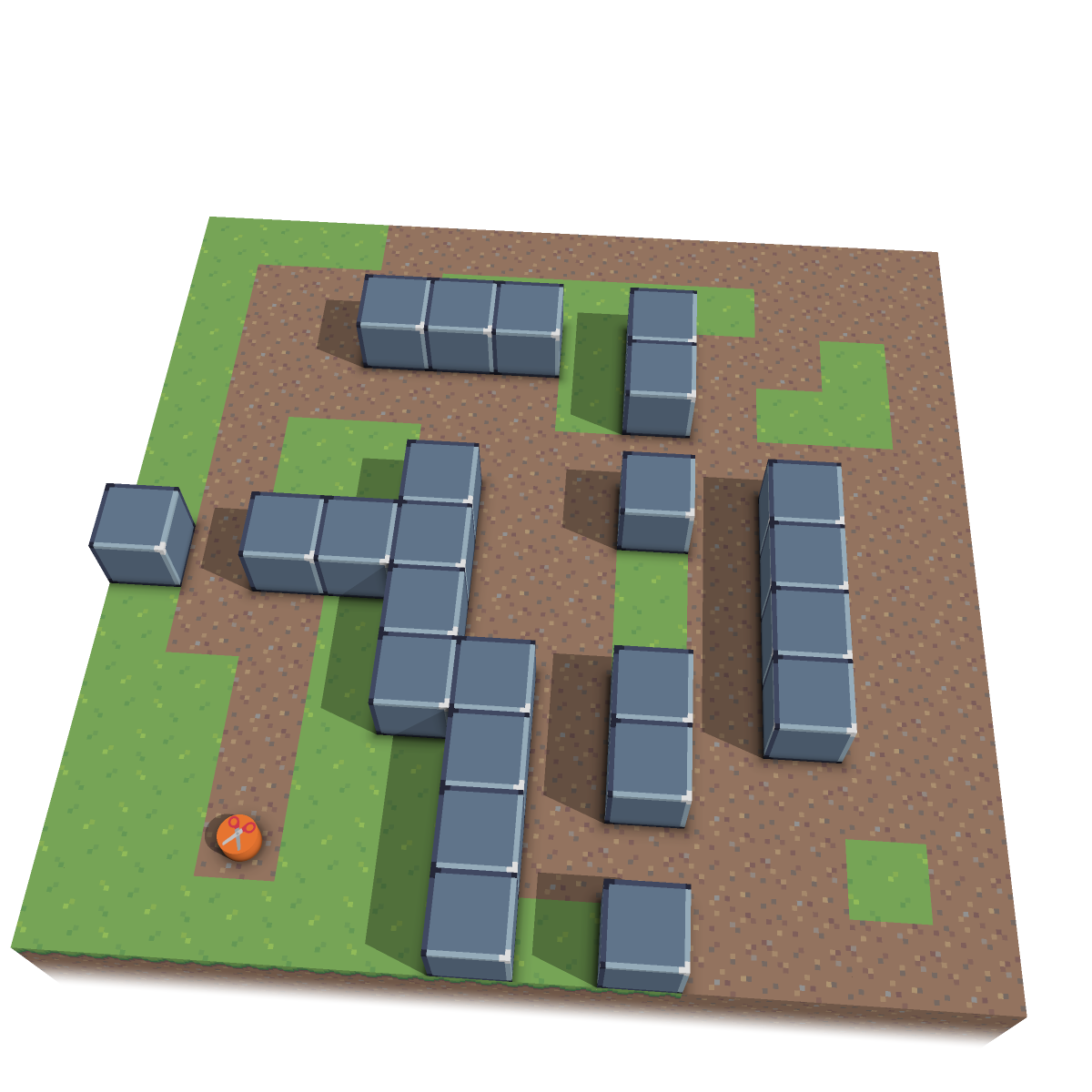}
         \\
         Dodgeball & Football & Hide & Lawnmower
         \\
         \includegraphics[width=0.24\linewidth]{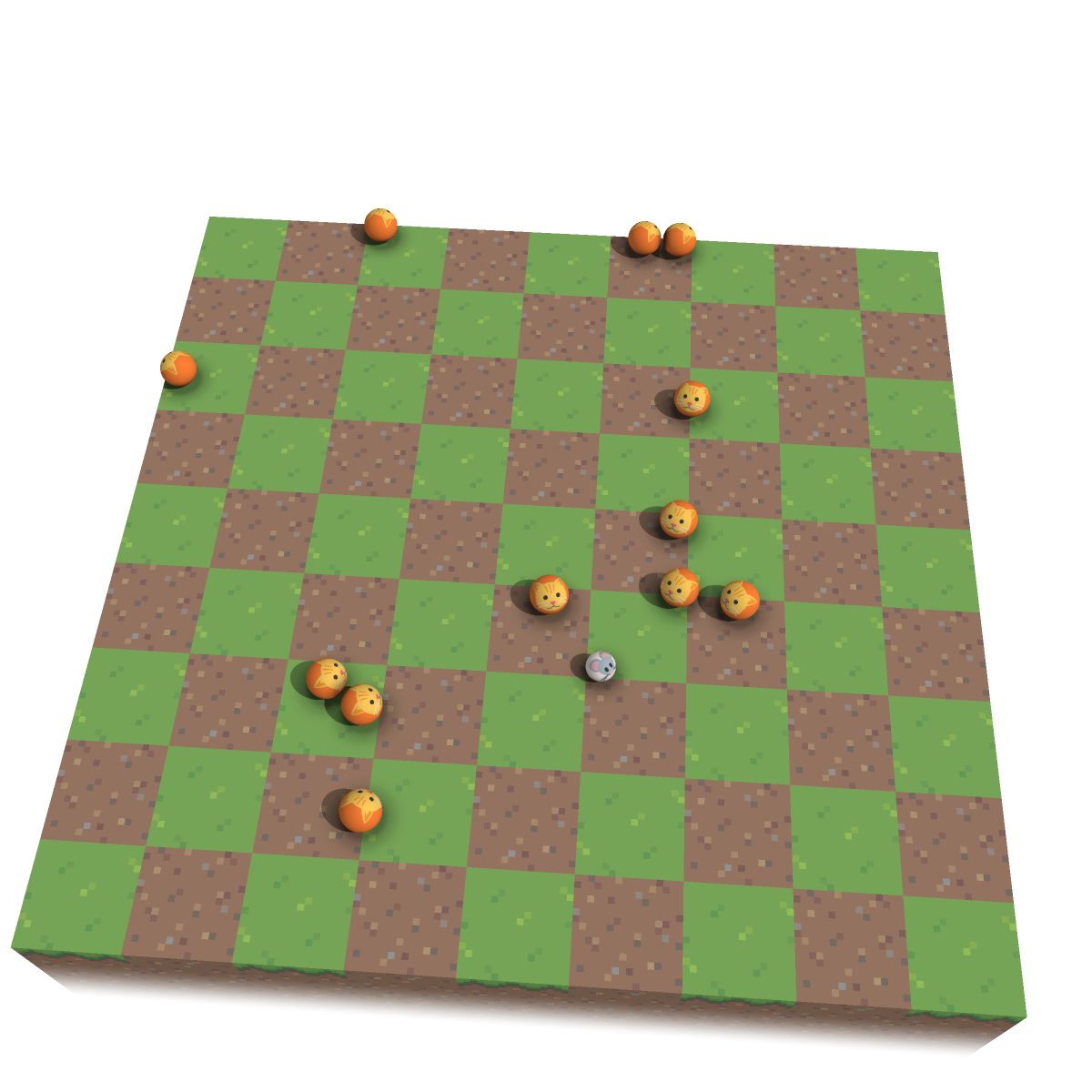}
         &
         \includegraphics[width=0.24\linewidth]{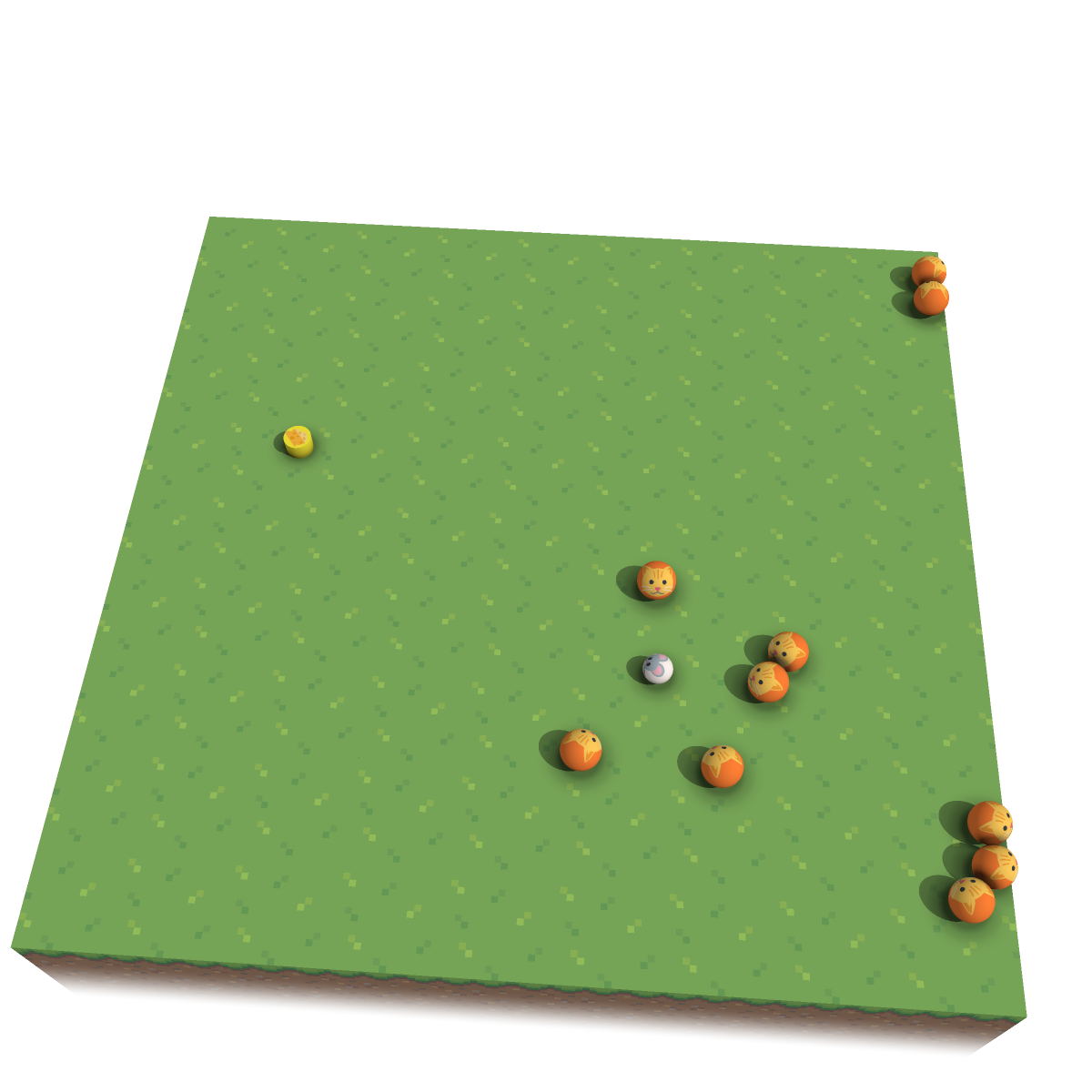}
         &
         \includegraphics[width=0.24\linewidth]{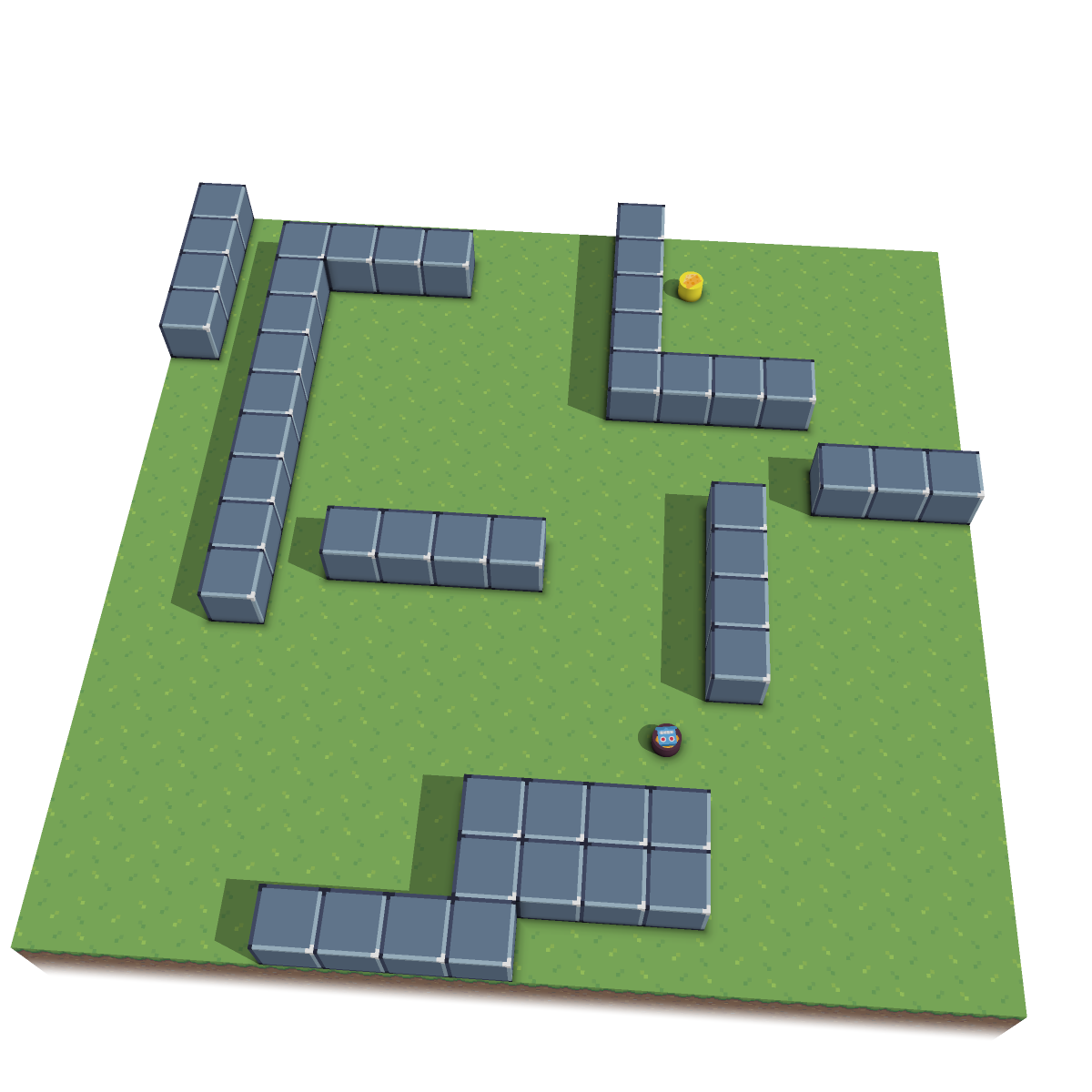}
         &
         \includegraphics[width=0.24\linewidth]{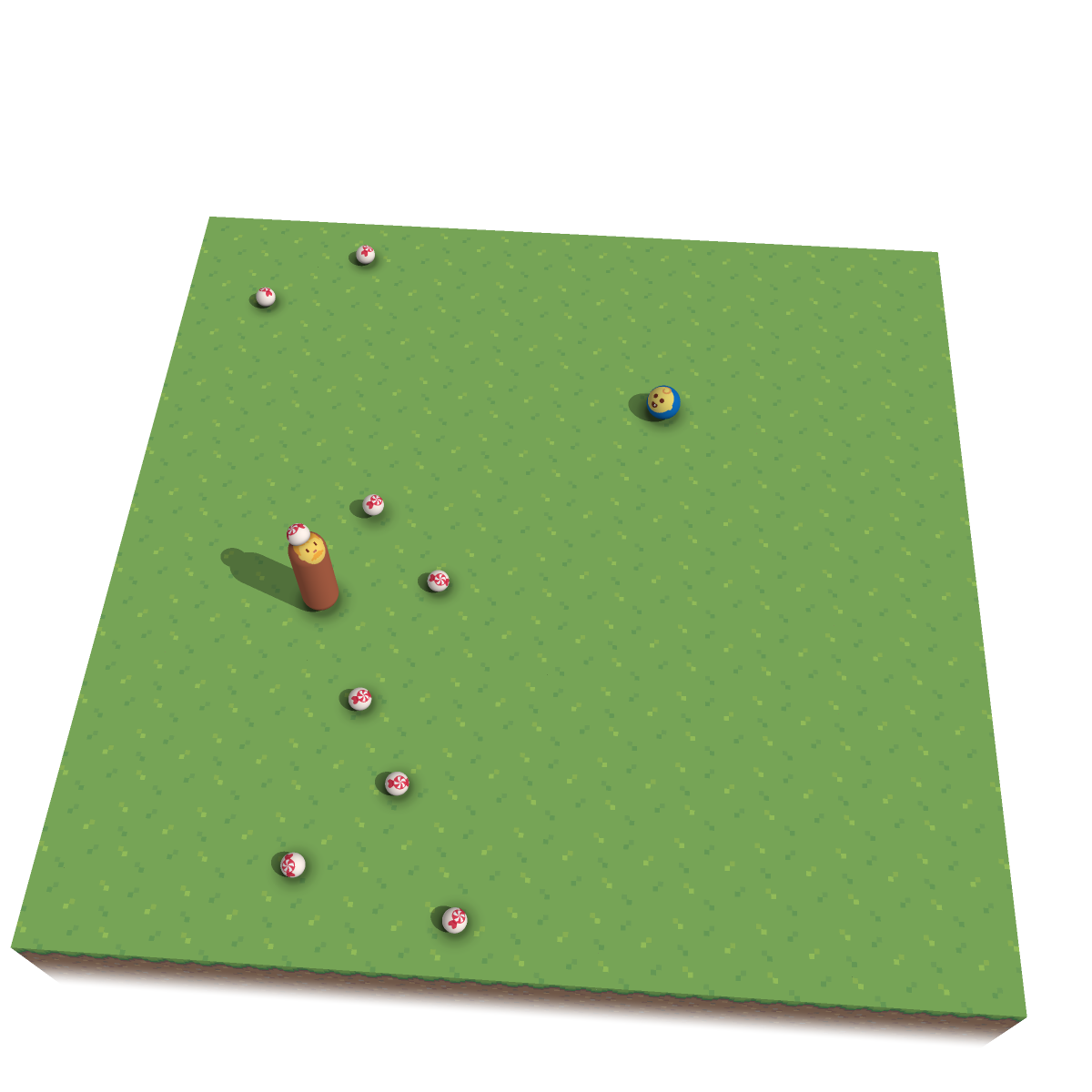}
         \\
         MouseCats & MouseCheeseCats & Pathfinding & ProtectTheChild
         \\
         &
         \includegraphics[width=0.24\linewidth]{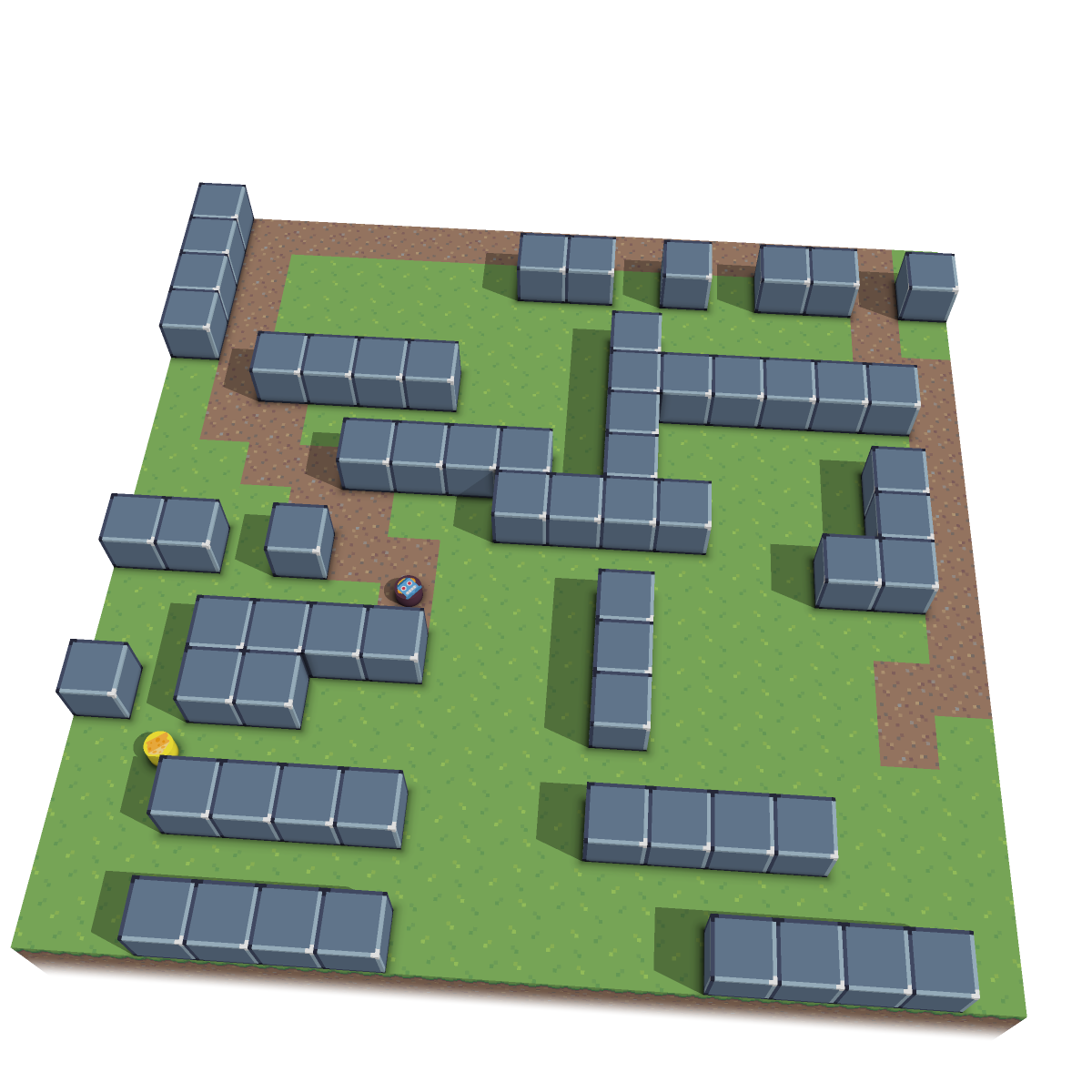}
         &
         \includegraphics[width=0.24\linewidth]{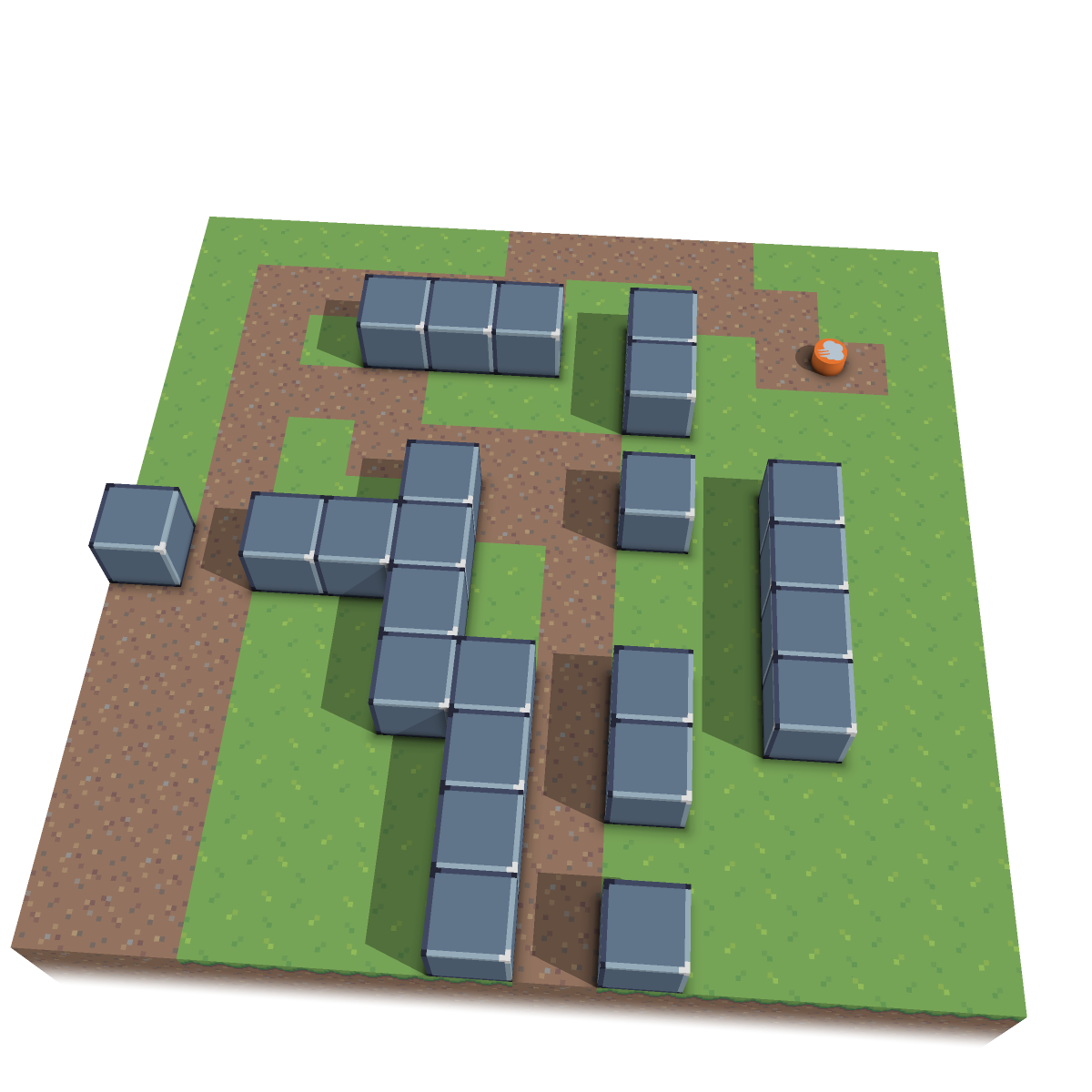}
         \\
         & Seek & Vacuum
    \end{tabular}
\caption{
Screenshots from all 14 games in our benchmark.
See the code repository for full descriptions of each game, including formal specifications in our game description language.
}
\Description{A grid of screenshots from the 14 benchmark games, showing their maps, characters, obstacles, goals, and interactive objects.}
\label{fig:benchmark_games}
\end{figure*}
\clearpage
\input{dsl-grammar}
\clearpage

\balance
\bibliographystyle{ACM-Reference-Format}
\nocite{SolarLezama2009Sketching}
\nocite{VermaEtAl2018PIRL}
\nocite{LiuEtAl2025LLMGuidedSearch}
\nocite{CarvalhoEtAl2024Reclaiming}
\nocite{InalaEtAl2020InductiveGeneralization}
\bibliography{references}
\end{document}

%% file: 1-Introduction.tex
\section{Introduction}

Autonomous behaviors are a crucial component of interactive virtual environments.
In games, they control the actions of non-player characters and allow the creation of playtesting agents that estimate game difficulty and expose bugs and other design problems.
In artificial intelligence and robotics, virtual environments are often used to train and test autonomous behaviors for embodied agents that must map observations to actions before deployment in the physical world.
Such behaviors can be implemented either through \emph{reactive} policies, which act based on immediate observations of the environment, or through \emph{planning}, which uses a model of the environment to anticipate the consequences of potential actions and chooses among them accordingly.
In this paper, we focus on reactive policies.

There are two common ways to author reactive policies. The first is to specify the behavior directly as a program, with finite-state machines and behavior trees as familiar special cases.
This approach underpins seminal computer graphics research on virtual character behavior~\citep{Boids,ArtificialFishes,MultiLevelCreatures}; it is both standard practice in commercial game AI and widely used in robot control, where modular and reactive behavior representations are valued~\citep{YannakakisTogelius2018AIGames,ColledanchiseOgren2018BTs,IovinoEtAl2022BTSurvey}.
The second approach is to derive behavior from goals or rewards. This goal-directed view is central to AI research: rational agents are commonly described as systems that perceive their environment and act to achieve their objectives, and intelligence has also been formalized as the ability to achieve goals across environments~\citep{RussellNorvig2021AIMA,LeggHutter2007UniversalIntelligence}. The most prominent modern example of this second approach is deep reinforcement learning (RL), where policies or value functions are represented by neural networks and optimized using reward signals~\citep{MnihEtAl2015DQN}.

Both high-level approaches have significant limitations. For the direct authoring approach, writing policies by hand requires effort and expertise, since the designer must translate an intended behavior into a reactive controller that handles edge cases, timing, partial observability, and adversarial behavior. Doing so is especially difficult in imperfect-information games, where fog of war forces agents to scout the map and reason from incomplete observations~\citep{VinyalsEtAl2017SC2LE}. A common shortcut is to give agents privileged access to state that a player-controlled character would have to infer from observation. Privileged state can make agents more competent, but it weakens believability when players notice that an agent is reacting to information it could not have observed. This downside motivates behavior synthesis methods that can produce effective policies using the same observations available to the character.

The reward-learning approach has complementary limitations. In deep RL, the learned policy is a neural network, so its behavior is difficult to inspect, edit, or constrain directly~\citep{GlanoisEtAl2024InterpretableRL}. This lack of interpretability makes artistic control hard: a designer cannot easily ask why the agent made a decision, change one local rule, or preserve a recognizable style after the game changes. It also limits portability across games, versions, and patches, since changes to dynamics, observations, or rewards can invalidate the learned policy and require new training or extensive retuning. Training is expensive because agents often need millions to hundreds of millions of simulator interactions~\citep{MnihEtAl2015DQN,HesselEtAl2018Rainbow}, with large-scale game agents requiring substantial distributed computation~\citep{BernerEtAl2019OpenAIFive,VinyalsEtAl2019AlphaStar}. Tools such as Unity ML-Agents make RL accessible to developers and researchers~\citep{JulianiEtAl2018Unity}, but the resulting policies remain a poor fit for everyday behavior authoring, where designers need cheap iteration, predictable control, editable artifacts, and fast runtime at deployment. The result is a gap between academic success on reward-driven learning and production practice, where hand-authored behaviors remain the dominant representation.

This paper studies a third point in the design space: automatically synthesizing \emph{programmatic} policies from rewards. The designer provides a reward function and a game simulator, and the system searches for a compact reactive program that scores well in the game. The resulting policy can be evaluated like an RL agent and inspected like a hand-written behavior script. This paradigm combines the main advantages of the two approaches above: reward-driven discovery without manual behavior coding and editable program artifacts instead of opaque neural policies. Because the output is a small program in  a domain-specific language (DSL), the method fits naturally into existing game development workflows, where designers already author, inspect, and revise behavior logic.

Prior work on programmatic policy search has shown some promising results, but existing evaluations concentrate on games set in small, discrete environments known as \emph{grid worlds}---Karel and MiniGrid are two prominent examples~\citep{CarvalhoEtAl2024Reclaiming,LiuEtAl2025LLMGuidedSearch}. While useful for controlled experiments, these benchmarks do not capture the structure of \emph{continuous} games. Many games enjoyed by players today require policies to reason about distances, directions, velocities, occlusion, contact, pursuit, collision avoidance, and physics over continuous space and time.
Existing programmatic policy search work has considered some continuous domains, such as racing and parking \citep{VermaEtAl2018PIRL,InalaEtAl2020InductiveGeneralization}, but these are isolated tasks with game-specific policy languages.
In contrast, we aim to study programmatic policy synthesis in a broader continuous-game setting, where many different behaviors for a wide range of games are expressed in a single DSL.

Our approach has two main components: a policy representation and a search algorithm for that representation.
For continuous games, the space of possible behaviors is also continuous, which makes direct search difficult. We address this challenge by designing a small DSL for reactive policies in continuous games.
The DSL discretizes the program search space into compact program expressions while still exposing the quantities that game characters need to act: distances, directions, contacts, and local geometric relationships. \emph{Raycasting} is the primary sensing primitive, allowing policies to query the surrounding scene in a form that is common in game engines.
The DSL also includes higher-order functions such as \texttt{maxDirection}, which let a policy choose directions which maximize some local geometric criterion. We designed the DSL to work well with program search, so its expressions are typed, compact, and biased toward reusable patterns of continuous-game behavior.

We design our policy search algorithm around two key ideas. First, we use \emph{antipatterns} to prune the search space. An antipattern is a banned expression schema, such as \texttt{(if (not b) A B)}, which is redundant with \texttt{(if b B A)}. We use dozens of such rules to eliminate equivalent, dominated, or unhelpful expressions permitted by our DSL's grammar. These rules reduce the number of programs that must be considered while preserving the same space of continuous behaviors. Second, we combine \emph{bottom-up} and \emph{top-down} program synthesis. Bottom-up search is performed by a symbolic enumerator that exhaustively generates and tests well-typed expressions of bounded size \citep{UdupaEtAl2013Transit,AlbarghouthiEtAl2013Escher,GulwaniEtAl2017ProgramSynthesis}. Top-down structure is provided by an LLM-based coding agent~\citep{OpenAICodex}, which writes a high-level policy stencil with holes. The coding agent can then call the bottom-up enumerator as a tool, asking it to fill a particular hole by testing candidate DSL expressions inside the simulator. We call this combined procedure \emph{agentic sketching}: the coding agent proposes the decomposition and global structure, while symbolic enumeration performs exhaustive local search over program fragments, following the sketching view of program synthesis as completing partial programs with holes \citep{SolarLezamaEtAl2006Sketch}.

We compare agentic sketching against pure bottom-up symbolic search and against the LLM-based coding agent alone.
Perhaps surprisingly, bottom-up enumeration outperforms the coding agent alone, despite running without an LLM on a single laptop CPU. This result suggests that exhaustive symbolic search is a strong baseline when the DSL is compact and carefully designed. The combined agentic sketching approach is substantially stronger than either of its constituent methods.
Agentic sketching uses the coding agent where it is most useful---to propose decompositions and high-level control structure---and uses symbolic enumeration where it is most useful: to search local DSL expressions exhaustively and cheaply.

Interestingly, our method also discovers effective reactive policies for games that would seem to require planning. For example, our Pathfinding game asks an agent to navigate around obstacles, but the DSL contains no graph search, waypoint planning, or A* primitive. The policy must act only through local geometric queries such as raycast distances. Despite this restriction, agentic sketching finds compact policies that navigate successfully, showing that rich continuous-game behaviors can emerge from reactive programs when the representation and search procedure are matched.

Our main contributions are:
\begin{itemize}
    \item A benchmark of 14 continuous games for programmatic policy search, together with a single unified DSL that can express policies for every game in the benchmark.
    \item A bottom-up symbolic search method that uses a set of do\-main-specific antipatterns to remove redundant and unhelpful expressions from the search space.
    \item \emph{Agentic sketching}, a program synthesis method that combines an LLM-based coding agent for high-level policy structure with exhaustive symbolic enumeration for local program fragments.
\end{itemize}

Together, these contributions make programmatic policy search a practical approach for authoring effective and interpretable behaviors in continuous games (Figure~\ref{fig:teaser}).
The code of our benchmark and algorithms is available at \url{https://github.com/mxgmn/BeyondTheGrid}.

%% file: 2-RelatedWork.tex
\section{Related Work}
\paragraph{Genetic Programming}
Genetic programming (GP) is a classical framework for fitness-driven program search: it initializes a population of random candidate programs, evaluates each program with a fitness function, and repeatedly forms new populations by selecting high-scoring individuals and applying variation operators such as mutation and crossover~\citep{Koza1992GP,PoliLangdonMcPhee2008FieldGuide}.
In the computer graphics literature, GP has been used to define the morphologies of virtual creatures~\citep{KarlSims1991,Sims1994EvolvingCreatures}, 
and to synthesize articulated motion~\citep{NgoMarks1993Spacetime,GritzHahn1995Articulated} and sensor--actuator controllers~\citep{VanDePanneFiume1993SAN}, while later work evolved behavior trees for game NPCs~\citep{PartlanEtAl2022EvolvingBehavior,LimEtAl2010DEFCON}. For continuous control, graph-based GP has produced small, interpretable policies~\citep{NadizarEtAl2024GraphGP, VacherEtAl2025MAPLE}. GP has also produced Atari policies that process raw pixel matrices~\citep{WilsonEtAl2018Atari} or quantized grids derived from pixels~\citep{KellyHeywood2017TPG}. These works, like ours, synthesize executable policies from performance feedback. Our focus is different: we introduce a set of games that share one object-level policy language and use it to compare symbolic enumeration with an LLM coding agent.

Our Lawnmower game is a continuous adaptation of Koza's Lawnmower problem~\citep{Koza1994GP2}, making GP a natural point of comparison. However, GP faces several difficulties in our setting. First, many of our tasks behave like needle-in-a-haystack problems: a random population of policies may receive uniformly zero reward, leaving selection with little useful signal to exploit. Second, standard mutation and crossover do not automatically respect our domain-specific antipatterns, which deliberately eliminate equivalent or behaviorally redundant programs. Third, many GP variants are prone to \emph{bloat}: evolutionary pressure can produce unnecessarily large programs through dead code, no-ops, and other redundancies (known as \emph{introns})~\citep{LangdonPoli1998Bloat,PoliLangdonMcPhee2008FieldGuide}. Bloated behavior programs are harder to read and edit, undermining one of the main reasons for synthesizing policies as programs in the first place. Our bottom-up enumeration method tests short programs first and thus avoids bloat.

\paragraph{``Decompiling'' neural networks}
Some prior methods learn programmatic policies by first learning an intermediate controller, then synthesizing a program that imitates it. PIRL trains a neural policy with deep reinforcement learning and uses it as an oracle for local search over programmatic policies \citep{VermaEtAl2018PIRL}. \citet{InalaEtAl2020InductiveGeneralization} follow a related teacher-student strategy, where an over-parameterized teacher is optimized first and a compact state-machine policy is trained to mimic it.
For such imitation learning to be successful, the DSL for programmatic policies must be sufficiently expressive to mimic the behavior of the teacher neural network (which can, if large enough, represent broad classes of continuous functions~\citep{Cybenko1989,Hornik1991}).
For example, in PIRL, the synthesized car-driving programs use symbolic expressions to switch between a set of parametrized PID controllers.
In contrast, our method is program-native and searches directly in the target policy language, without a neural policy as an intermediate representation. It therefore applies even when this language is intentionally restrictive and cannot approximate arbitrary neural network policies.

\paragraph{Modern programmatic policy search}
Recent work has produced a range of methods for searching over programmatic policies, including direct search in the program space, search in learned latent spaces, and LLM-guided search \citep{CarvalhoEtAl2024Reclaiming,LiuEtAl2025LLMGuidedSearch}. These methods demonstrate that compact programs can be effective policy representations, but they focus on small, discrete grid-world domains such as Karel, Karel-Hard, and MiniGrid.
A small number of prior works have considered continuous control domains: \citet{VermaEtAl2018PIRL} evaluate on TORCS racing, and \citet{InalaEtAl2020InductiveGeneralization} study a parking task. However, these are single-domain evaluations with task-specific policy languages. Our benchmark contains 14 continuous games whose policies are all expressed in a single DSL, allowing us to study programmatic policy search across a broader set of continuous-space character behavior problems.

\paragraph{Deep RL}
Deep reinforcement learning is an established method for deriving behavior from reward: it represents a policy or value function with a neural network and optimizes that network to maximize reward~\citep{MnihEtAl2015DQN}. Its success has led to broad evaluation across games and multi-agent simulations. A large evaluation of PPO-based multi-agent methods~\citep{YuEtAl2022MAPPO} spans several complementary testbeds. MPE~\citep{LoweEtAl2017MADDPG} includes predator--prey and keep-away; these tasks resemble Chase, MouseCats, MouseCheeseCats, ButtonCats, ButtonCheeseCats, and Hide in our benchmark. Our movement model is more complex because it also includes collisions between objects. SMAC~\citep{SamvelyanEtAl2019SMAC} studies cooperative combat under partial observability, Google Research Football~\citep{KurachEtAl2020Football} is directly comparable to our Football game, and the Hanabi Learning Environment~\citep{BardEtAl2020Hanabi} studies cooperation under imperfect information. Unity ML-Agents~\citep{JulianiEtAl2018Unity} provides further close analogues: Hallway uses local raycasts for navigation, like Pathfinding and Seek; Pyramids requires a switch to spawn a later target, like ButtonCheese; Food Collector resembles ProtectTheChild's competition over objects; and its DodgeBall showcase directly parallels Dodgeball. These benchmarks show that the individual behaviors in our suite are not new. In the cited studies, behavior is encoded in neural-network weights, making it hard for a designer to understand a decision, change one rule, or impose a behavioral constraint~\citep{GlanoisEtAl2024InterpretableRL}. Our benchmark retains reward-driven discovery but requires all 14 games to use the same policy DSL, so the result of search is a small program that can be inspected and edited directly.

Deep Symbolic Policy~\citep{LandajuelaEtAl2021SymbolicPolicies} is an unusual exception: it treats the generation of a symbolic controller as an RL problem. A recurrent neural network samples mathematical expressions and is rewarded according to how well each expression controls the system. The next subsection considers a more recent version of the same broad idea: pretrained language models that write and revise executable policies from feedback.

\paragraph{LLMs for programmatic policies}
Several recent systems use language models to construct or revise behavior programs and action plans. Weng's Heuristic Learning~\citep{Weng2026LearningBeyondGradients} uses a coding agent to revise rule-based controllers using rewards, logs, tests, and replays, and reports scores competitive with PPO baselines. This resembles our coding-agent baseline, which also writes and tests complete policies. Weng's controllers are written in Python. In our benchmark, every controller must use the same restricted policy language; this lets us compare the coding agent with exhaustive program search on the same problems.

PORTAL~\citep{XuEtAl2025PORTAL} shares our use of one policy language across games: it generates behavior trees for thousands of 3D games and revises them using game scores and feedback from a vision-language model. However, the paper does not release its game suite as a reusable public benchmark.

Voyager~\citep{WangEtAl2024Voyager} explores Minecraft under an automatically generated curriculum, iteratively writes and debugs code using execution feedback, and stores successful programs in a reusable skill library. Voyager demonstrates this process within Minecraft rather than providing a cross-game benchmark for policy synthesis. Inner Monologue~\citep{HuangEtAl2023InnerMonologue} uses an LLM to revise high-level robot plans using success signals, scene descriptions, and human feedback. It selects and sequences predefined robot skills rather than synthesizing new controller programs.

%% file: 3-Benchmark.tex
\section{Benchmark}

In designing a benchmark for evaluating different character behavior synthesis methods in continuous games, we need to go beyond grid worlds, where actions are discrete moves between grid cells, and geometry is reduced to cell adjacency.
We therefore need a simulation engine and a set of games within it that expose the geometric and dynamical phenomena of real-time character control.
Specifically, the benchmark should contain scenarios that require behavior policies to handle distances, directions, velocities, occlusion, contact, collision avoidance, pursuit, evasion, navigation, and interactions with dynamic objects.

Figure~\ref{fig:benchmark_games} shows screenshots of all of the games in our benchmark; full formal specifications of each game in our game description language are provided in the code repository.
The games include coverage mechanics (Lawnmower, a continuous adaptation of Koza's classic problem~\citep{Koza1994GP2}, and Vacuum), pursuit mechanics (Chase), avoidance and mixed-objective mechanics (MouseCats, MouseCheeseCats, ButtonCats, and ButtonCheeseCats), navigation mechanics (Pathfinding and Seek), line-of-sight mechanics (Hide), object interaction mechanics (ButtonCheese, ProtectTheChild), and the use of basic rigid body physics (Dodgeball, Football).

\subsection{Game Simulation Engine}
Our games are composed of objects, and each object has a primitive shape: a sphere, cylinder, capsule, or cuboid of varying size. Each object also has a position, a yaw direction represented as a 2D unit vector in the ground plane, and a 3D velocity vector. This common representation allows the benchmark to cover a variety of tasks while keeping the underlying sensing and control problem consistent across games.

For the same reason, all games also use the same action interface. At each simulation step, the controlled character outputs an action in $S^1 \times \{0,1\} \times \{0,1\}$, where the unit circle $S^1$ specifies a target direction, the first Boolean specifies whether the \texttt{forward} button is pressed, and the second Boolean specifies whether the \texttt{use} button is pressed. The character movement model is determined by the target direction and the \texttt{forward} button. The character has a current facing direction and a finite turn rate. If the \texttt{forward} button is not pressed, the character rotates in place toward the target direction along the shorter of the two possible arcs. If the \texttt{forward} button is pressed, the character rotates toward the target direction while moving forward along its current facing direction at fixed speed. This nonholonomic control scheme gives characters bounded-curvature motion related to the Dubins car model \citep{Dubins1957}. The finite turn rate makes movement more physically realistic: pursuing, kiting, dodging, and navigating around obstacles all depend on facing direction and cannot be achieved by choosing an instantaneous velocity vector. Finally, the \texttt{use} button triggers game-specific interactions, such as picking up, throwing, or dropping held objects.

\subsection{Evaluation Protocol}
We use our benchmark to evaluate whether programmatic policies produced by different search methods \emph{generalize}, that is, how well a policy performs in different environments (or ``levels'') within the same game.
To aid in this assessment, we use randomized procedural generation: the geometry of stationary features such as walls, as well as the initial positions of agents and other interactive objects, is determined based on a random seed.
In designing procedural level generators for each game, we take care to ensure that they cannot produce any pathological or ``unwinnable'' scenarios, e.g. environments where the character starts out trapped in some corner of the scene.

We evaluate a candidate policy by rolling it out in the simulator for $N$ different random seeds.
The score of the policy is then taken as the minimum score across all $N$ of these runs.
We use the minimum as our aggregation function for two reasons.
First, it is robust to outliers, which can arise when a random seed happens to produce a fortuitous initial configuration where the character achieves a very high score.
Second, min aggregation allows for more efficient search: if after running on, say, 2 random seeds, a candidate policy already has a worse minimum score than the best policy found so far, there is no need to run that policy on the remaining $N - 2$ random seeds.

%% file: 4-DSL.tex
\section{Domain-Specific Language (DSL)}

To design and evaluate methods that search for programmatic behavior policies, we must first specify the language in which those policies will be expressed.
In crafting a domain-specific language for behavior policies, we seek to satisfy the following design goals:
\begin{enumerate}
    \item \textbf{Unified Language}: one DSL should be able to specify behaviors for all games in our benchmark.
    \item \textbf{Incomplete Information}: the DSL should support games where the agent has incomplete information about the state of the world, which is a common setup in video games and robotics.
    \item \textbf{Portability}: programs in the DSL should be easy to port to any game engine or robotic simulation framework, to allow other researchers and practitioners to use our methods.
    \item \textbf{Efficient Search}: the DSL should support efficient program search by allowing interesting, complex behaviors to be expressed by compact programs.
\end{enumerate}

With these design goals in mind, we made the following decisions in designing our policy DSL:

\paragraph*{Separate game description language}
In our system, a separate game description language (documented in the code repository) specifies game level layout, game dynamics, and how the character receives rewards, while the policy DSL specifies how characters act within these games.
This design decision supports \textbf{(1)} \textbf{Unified Language}, as the single \texttt{use} action exposed in the policy DSL can trigger a variety of game dynamics by invoking an \texttt{onUse} handler for the object specified by the game description language.
It also supports \textbf{(3)} \textbf{Portability}, as this factored design makes it easier to substitute in other game description formats used by other game engines.

\paragraph*{Simple syntax and grammar}
Our policy DSL is written in a simple, LISP-like S-expression syntax.
It has a small number of types: \texttt{void}, \texttt{bool}, \texttt{float}, \texttt{vector}, \texttt{object}, \texttt{action}, \texttt{list}, and \texttt{tile}.
It also has a simple grammar, consisting primarily of arithmetic operations, (in)equality operations, conditionals, and function calls.
Table~\ref{tab:dsl-grammar} gives the grammar in a machine-readable form.
Keeping the language simple supports \textbf{(3)} \textbf{Portability}; it also supports \textbf{(4)} \textbf{Efficient Search}, as simpler languages allow fewer possible programs than more complex ones.

\paragraph*{Sensors}
Rather than having access to complete game state, the policy program observes the game world through \emph{sensors} (\textbf{(2)} \textbf{Incomplete Information}).
A small number of atomic sensors are built into our policy DSL (e.g. \texttt{raydist}); these sensors are implemented using raycasting, which is a commonly available primitive in game engines and robot simulation frameworks (\textbf{(3)} \textbf{Portability}).
These kinds of sensors are inspired by the minimal robots of \citet{LaValle2012SensingFiltering}: in contrast to high-dimensional sensors such as images (as commonly used in deep RL), these low-dimensional sensors lead to fewer distinct values that can appear in programs, helping to reduce the search space (\textbf{(4)} \textbf{Efficient Search}).

\paragraph*{Higher-order functions}
Our policy DSL also includes higher-order functions that make use of lambda expressions.
For brevity, we write the lambda expression which would traditionally be expressed as \texttt{(lambda (x) (f x))} as simply \texttt{(f \#0)}.
These functions allow expressing complex patterns compactly (\textbf{(4)} \textbf{Efficient Search}). For example, \texttt{(maxDirection (raydist \#0))} returns the direction from the character in which a ray travels farthest before colliding with something else in the scene.

\begin{figure}[t!]
    \centering
    \includegraphics[width=\linewidth]{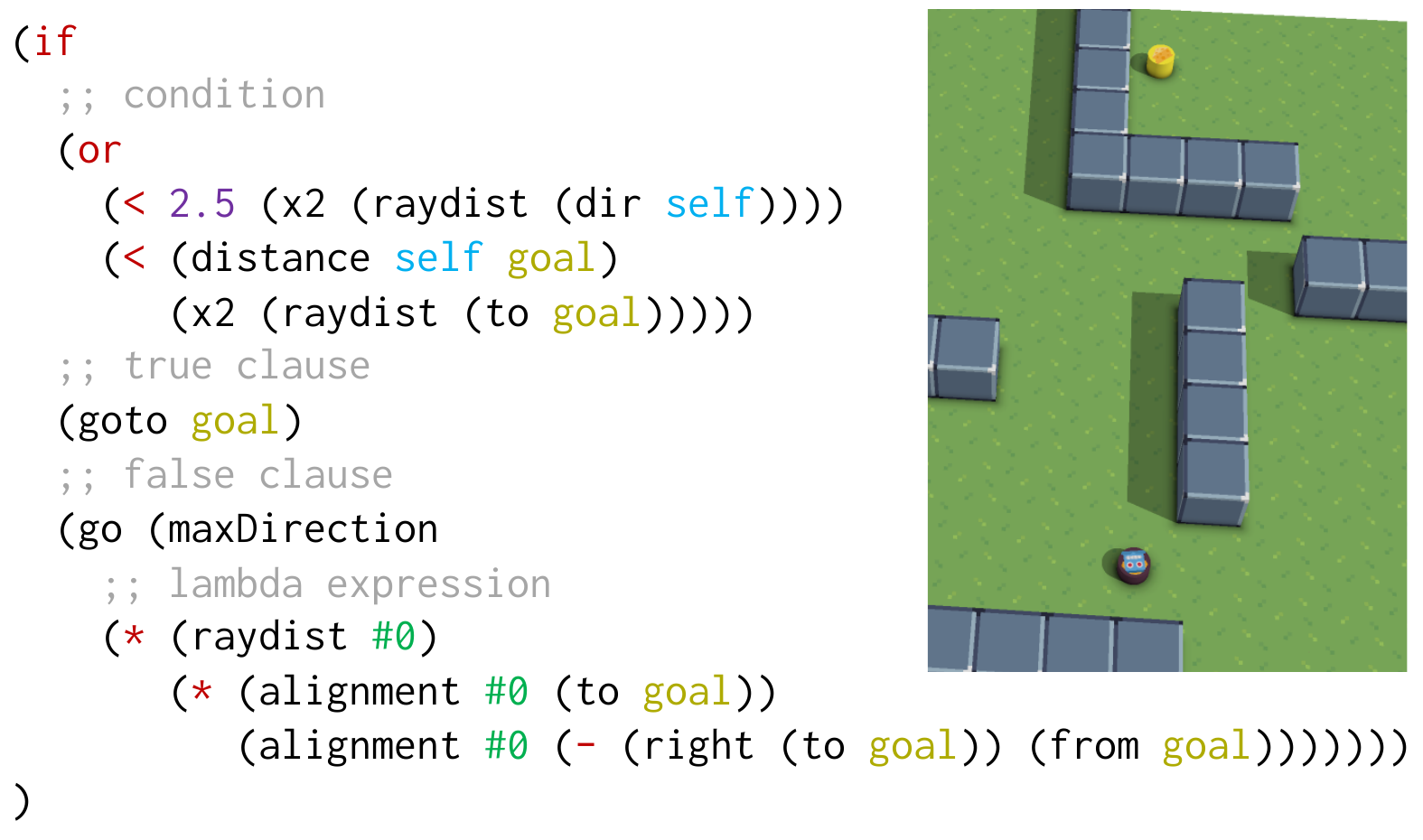}
    \caption{
    An example program in our policy DSL for the Pathfinding game (right).
    The policy uses a conditional branch to decide whether the agent (the blue robot) should proceed directly toward the goal (yellow cheese) or use more complex logic to find a way around obstacles.
    The \texttt{or} function shown here is not part of the DSL; it is equivalent to nested \texttt{if} expressions and is used here for readability.
    In the false branch, the agent chooses the direction that maximizes the product of three quantities: the raycast length in that direction, the direction's alignment with the goal, and its alignment with the vector \texttt{(- (right (to goal)) (from goal))}.
    This policy was discovered by our agentic sketching algorithm.
    }
    \Description{A Pathfinding game scene beside a synthesized policy program that conditionally moves toward the goal or follows raycast-based obstacle-avoidance logic.}
    \label{fig:dsl_example}
\end{figure}

%% file: 5-Algorithm.tex
\section{Policy Search Algorithm}
Our algorithmic contribution consists of two parts. First, we improve bottom-up symbolic search using DSL-specific \emph{antipatterns}. Second, we combine symbolic search with a coding agent through \emph{agentic sketching}.

\subsection{Improved bottom-up search}
We start from a standard bottom-up enumerative synthesis algorithm, which constructs expressions from a grammar by first adding terminals and then repeatedly applying grammar productions to expressions that have already been constructed \citep{AlbarghouthiEtAl2013Escher,UdupaEtAl2013Transit,GulwaniEtAl2017ProgramSynthesis}. Since complete expressions are produced during search, each candidate policy can be executed directly in the simulator.

The main challenge is the size of the expression space. Even for the toy grammar $E::=x\mid f(E,E)$, the number of expressions $N(d)$ of depth at most $d$ satisfies $N(d) = 1 + N(d-1)^2$, so the number of candidates grows doubly exponentially with depth. Our DSL is substantially larger than this toy grammar, so an unconstrained enumeration becomes infeasible even for short programs. Types provide the first reduction: the DSL is typed, and the enumerator constructs only well-typed expressions.

A standard pruning technique in bottom-up synthesis is \emph{observational equivalence}: if two expressions evaluate to the same values on all provided inputs, the synthesizer keeps only one representative, since replacing one expression with the other preserves behavior on the examples \citep{AlbarghouthiEtAl2013Escher,UdupaEtAl2013Transit}. This technique is highly effective in programming-by-example, where the specification provides a fixed and usually small set of inputs. Policy search has a different structure. The inputs to a policy are simulator states, and these states are generated by the actions taken during rollout. There is no fixed example set that defines the problem. 
We therefore use syntactic \emph{antipatterns} instead of observational equivalence. An antipattern is an expression schema that matches expressions we choose to exclude from the search. During bottom-up enumeration, any expression that matches an antipattern is discarded and never added to the candidate set. This pruning is valuable because removing one expression also removes every larger expression that would have used it as a subexpression.

Some antipatterns eliminate exact redundancies. For example, we exclude expressions of the form \texttt{(not (not A))}, since they are equivalent to the shorter expression \texttt{A}. We also exclude schemas such as \texttt{(if (not B) A C)}, which can be written as \texttt{(if B C A)}, nested directional inverses such as \texttt{(left (right A))}, expressions such as \texttt{(== self none)}, and \texttt{(to self)}, which denotes a zero vector by the semantics of the DSL. Another example is \texttt{(maxDirection (alignment \#0 V))}: the direction that maximizes alignment with a vector \texttt{V} is \texttt{V} itself. Antipatterns can also be context-sensitive. For example, after entering a branch guarded by condition \texttt{B}, we exclude nested tests that branch on \texttt{B} again, since the condition value is already known in that context. 
Other antipatterns are design choices rather than exact equivalences. For example, we exclude expressions such as \texttt{(distance (equipped A) B)}. This expression is not always equivalent to \texttt{(distance A B)}, but in our domains it is usually enough to measure distances between base objects rather than between their equipped objects. Such antipatterns trade a small amount of expressivity for a much smaller search space. This tradeoff is appropriate for our setting because the goal is to cover the space of useful continuous behaviors efficiently, not to enumerate every syntactically possible expression. The code repository lists the full antipattern set.

The effect of antipatterns can be seen even in the toy grammar above. If we add the antipattern $f(\cdot, f(\cdot,\cdot))$, then the number of expressions of depth at most $d$ becomes linear. This example is extreme, but it illustrates why local syntactic exclusions can produce large reductions in bottom-up enumeration.

After enumerating expressions up to a fixed size bound, we select all expressions of type \textsc{Action} as candidate policies. Before simulation, we apply a simple static filter to remove policies that cannot affect the environment in the required way. For example, a policy that contains no movement operator such as \texttt{go} or \texttt{goto} is discarded. In games with an \texttt{onUse} trigger, policies that never call \texttt{use} are also discarded. The remaining candidates are evaluated by rolling them out in the simulator.

More sophisticated search procedures could bias enumeration toward expressions that are predicted to score well, as in cost-guided bottom-up synthesis \citep{AmeenLelis2023BeeSearch}. We deliberately use exhaustive enumeration over the pruned space. This keeps the method simple and makes the contribution of the DSL and antipatterns easy to measure. 
\subsection{Agentic sketching}
Our method gives an off-the-shelf coding agent, such as OpenAI Codex~\citep{OpenAI2025Codex}, access to the symbolic enumerator described above. The agent writes and revises the high-level policy structure and invokes the enumerator as a tool. We call the resulting procedure \emph{agentic sketching}: the coding agent proposes the decomposition and global structure, while symbolic enumeration performs exhaustive local search over program fragments, following the sketching view of synthesis as completing partial programs with holes~\citep{SolarLezamaEtAl2006Sketch}.

The central object in agentic sketching is a policy sketch. A sketch is a DSL program with a single typed \textsc{Hole} node. Sketches are incomplete programs, so they cannot be evaluated directly. To use the search tool, the coding agent specifies three pieces of information. First, it provides the policy sketch $S[\textsc{Hole}]$, whose surrounding context determines the required type of the missing expression. Second, it gives a maximum size bound for the expressions to be searched. Third, it may provide sketch-specific antipatterns that rule out expressions that are redundant or unhelpful in this particular context.

The sketch-specific antipatterns let the agent use information from the surrounding program. For example, consider the sketch
\[
    \texttt{(if empty (if amAttacker (goto target) HOLE) use)}.
\]
The hole is reached only in the branch where \texttt{empty} and \texttt{amAttacker} are already known. In this context, candidate subexpressions that test these conditions again enlarge the local search space without adding useful behavior. The agent can therefore pass antipatterns that exclude such tests from the hole-filling search. These local antipatterns play the same role as the global antipatterns from the previous subsection, but they are specialized to the current sketch.

Given a sketch, a size bound, and optional local antipatterns, the symbolic tool enumerates all well-typed expressions of the hole type up to the requested size. For each expression $e$, it substitutes $e$ into the sketch to form the complete policy $P[e]$, simulates that policy in the game, and records its reward. The tool then returns the best-scoring subexpressions at each size, together with their completed policies and scores. The coding agent uses this feedback to decide whether to keep the current sketch, revise the surrounding structure, increase the search bound, add new antipatterns, or try a different decomposition. In this way, the agent performs top-down search over policy sketches, while the enumerator performs bottom-up search inside each selected hole.

%% file: 6-EvaluationAndResults.tex
\section{Evaluation and Results}
We evaluate three methods: pure bottom-up symbolic search, a pure coding agent, and agentic sketching. In the two conditions that use a coding agent, we use OpenAI Codex with GPT-5.5 and Extra High reasoning \citep{OpenAI2025Codex}. Codex is given a sandbox folder containing the game description, the DSL description, and the executable environment. The environment accepts policies through command-line arguments, so Codex can test a candidate policy by running the executable with that policy as input. Agentic sketching uses the same interface: calls to the symbolic search tool are implemented as command-line calls that specify a sketch, a hole type, a search bound, and optional local antipatterns. The full coding agent prompts are included in the code repository.

Each method is given a 20-minute wall-clock budget per game. At the end of the budget, the method is terminated and the best policy found so far is recorded. We evaluate policies on 5--50 random seeds, depending on the structure of the game. Games with a single uninterrupted episode, such as MouseCats, are evaluated on 50 seeds. Games that chain many short subepisodes together, such as Pathfinding, where a new target spawns after the agent reaches the current target, are evaluated on 5 seeds. All games except Football run for one minute at 60 frames per second, for a total of 3600 simulation steps. Football runs for three minutes because it uses a larger map and characters need more time to traverse it.

\begin{figure*}
  \centering
  \includegraphics[width=\textwidth]{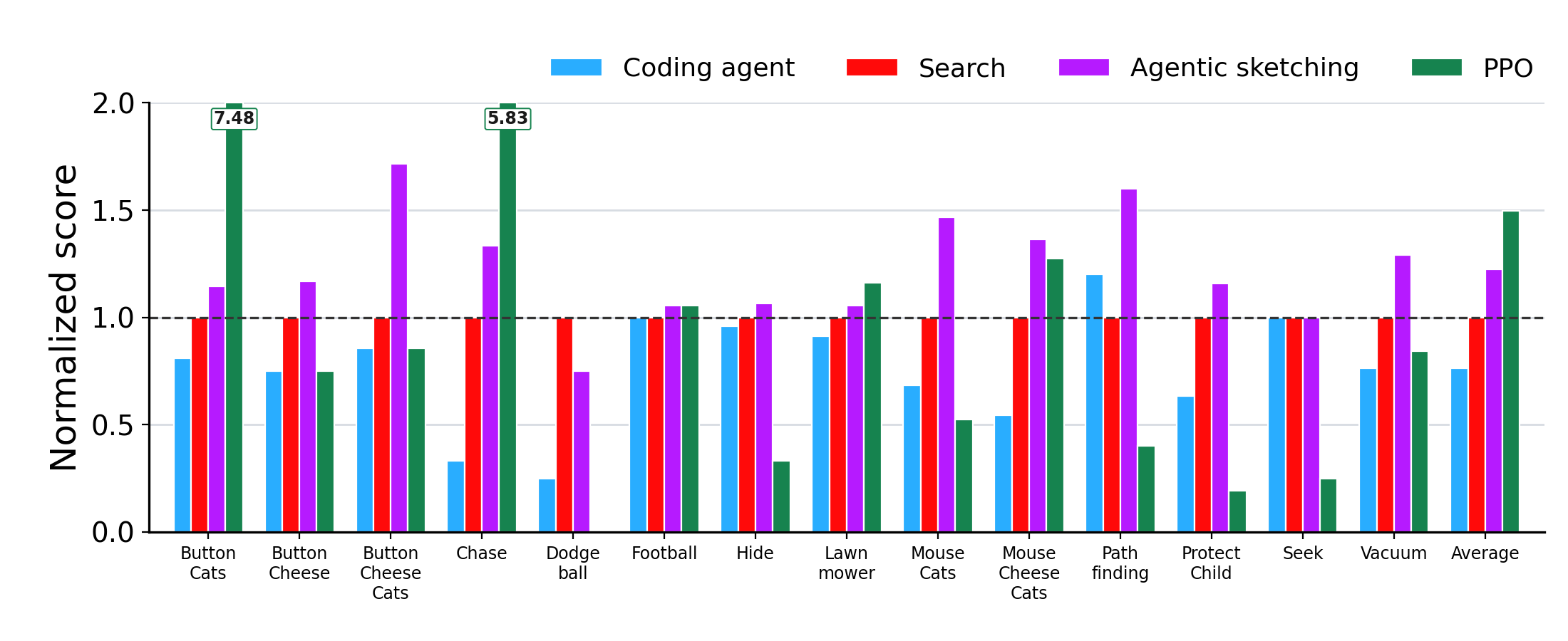}
  \caption{Best scores normalized per game relative to Search. Averages are computed after per-game normalization.}
  \Description{A grouped bar chart compares normalized scores for Coding agent, Search, Agentic sketching, and the best PPO minimum observed across all checkpoints for each benchmark game and their average.}
  \label{fig:normalized-results}
\end{figure*}

Figure~\ref{fig:normalized-results} shows the normalized benchmark results. We find that pure bottom-up search substantially outperforms the pure coding agent, while running without an LLM on a single laptop CPU (AMD Ryzen AI 9 HX 370 with all cores utilized). This result is notable because the search method is deliberately simple: it enumerates all well-typed expressions up to a size bound, removes expressions that match antipatterns, and tests the remaining candidate policies in the simulator. The coding agent performs far fewer policy evaluations and often commits to a narrower search strategy. In a post-hoc diagnostic, we gave Codex the best policy found by symbolic search and asked it to test the policy and explain why it missed it. In ButtonCheese, for example, search found a policy that repeatedly presses the button by orbiting it, waits until many cheese objects have spawned, and only then switches to a cheese-collection run (Figure~\ref{fig:teaser}). This behavior scores well because it avoids wasting time traveling back and forth between the button and individual cheese objects. The coding agent's explanations usually attributed the miss to an overly narrow framing of the task and insufficient exploration. This failure mode is consistent with recent evidence that LLMs still struggle with spatial and geometric reasoning in structured domains \citep{LiEtAl2024StepGame,LuoEtAl2026GeoGramBench}.

Agentic sketching outperforms both pure methods. The coding agent contributes high-level structure and decompositions that pure enumeration may not reach within the time budget, while the symbolic tool fills local holes by exhaustively testing many DSL expressions. This division of labor is visible in the aggregate results in Table~\ref{tab:aggregate-results}. As expected, the pure coding agent performs the fewest policy tests. Agentic sketching performs fewer tests than pure search, because it splits its budget between LLM inference and symbolic enumeration.
On average, agentic sketching spends 91\% of its time on LLM processing, 8.2\% of its time on policy testing, and 0.8\% on expression enumeration, including the elimination of expressions with antipatterns.
Table~\ref{tab:agentic-tool-calls} summarizes the agentic-sketching tool calls. Table~\ref{tab:expression-sizes} reports the maximum expression sizes used by each method.
Despite testing an order of magnitude fewer policies than pure search, agentic sketching achieves the highest average normalized score.
We also tested an ablated version of agentic sketching in which the agents cannot use sketch-specific antipatterns. Table~\ref{tab:ppo-results} shows that these antipatterns improve scores in 8 games, while scores remain unchanged in the other 6.

\begin{figure}[t]
  \centering
  \includegraphics[width=\columnwidth]{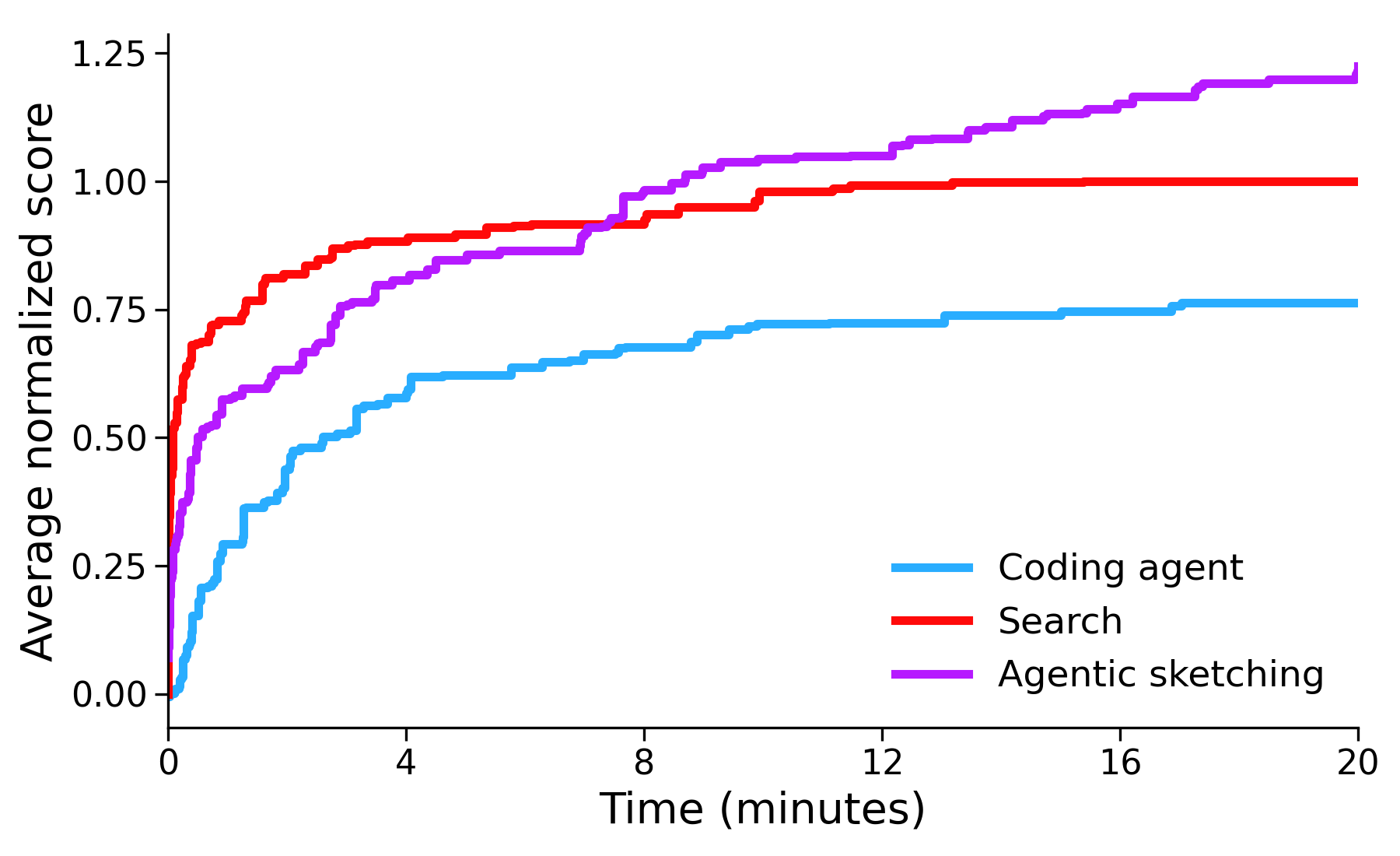}
  \caption{Average normalized score of the three policy search methods over the full 20-minute search interval. Search initially outperforms agentic sketching, but agentic sketching eventually overtakes it; both outperform the coding agent.}
  \Description{Line graph of average normalized score over time for Coding agent, Search, and Agentic sketching.}
  \label{fig:average-normalized-progress}
\end{figure}

Figure~\ref{fig:average-normalized-progress} shows average normalized score over the 20-minute budget; Figure~\ref{fig:search-history} shows the search progress for each game. Symbolic search improves fastest at the beginning, which reflects the efficiency of exhaustive enumeration over small expressions. Its progress then slows and it is overtaken by agentic sketching. This pattern suggests that bottom-up enumeration is strong for cheap local search, while coding agents are useful for proposing the larger program structures that make local search effective.

On average, policy search required 1.07 rollouts per policy. Early termination under the minimum-score objective keeps this number low: once a candidate policy has performed poorly enough on one seed, the remaining seeds often need not be evaluated. Sorting seeds and testing the hardest ones first further reduces the number of rollouts.

\begin{table}
  \centering
  \begin{tabular}{lrr}
    \toprule
    Method & Normalized score & Policy tests \\
    \midrule
    Coding agent & 0.76 & $5.39 \times 10^2$ \\
    Search & 1.00 & $1.99 \times 10^6$ \\
    Agentic sketching & 1.23 & $1.71 \times 10^5$ \\
    \bottomrule
  \end{tabular}
  \caption{Aggregate normalized score and average number of policy tests across benchmark games.}
  \label{tab:aggregate-results}
\end{table}

\subsection{Deep reinforcement learning baseline}
We additionally trained neural policies with proximal policy optimization (PPO)~\citep{SchulmanEtAl2017PPO}. We used the action-masked MaskablePPO implementation from Stable-Baselines3 2.9.0~\citep{RaffinEtAl2021SB3}; each game was exposed through Gymnasium 1.3.0 vector-environment interfaces~\citep{TowersEtAl2024Gymnasium}. For each game, we tuned PPO hyperparameters using successive halving~\citep{JamiesonTalwalkar2016SuccessiveHalving}. All games were trained for 100 million simulator frames. Figure~\ref{fig:ppo-search-history} shows the learning curves. Across all 14 games, the average wall-clock time was 140 minutes per $10^8$ training frames.

\begin{sloppypar}
Table~\ref{tab:ppo-results} and Figure~\ref{fig:normalized-results} compare the best policies returned by each method. PPO outperforms the program synthesis approaches on average, but its advantage is driven mostly by the ButtonCats and Chase results. Inspection of the playouts shows that the ButtonCats PPO agent learned to circle the button and press it rapidly. Programmatic agents discovered the same behavior, but the DSL prevents them from pressing the button as frequently. Similarly, a Chase PPO agent learned to collide with the target multiple times in a short period.

Among the games where PPO was unsuccessful, Dodgeball requires the agent to pick up a ball, throw it, and hit a moving target before receiving its first reward. In ProtectTheChild, a positive reward is added each frame, but most of the difference in the final score arises near the end of the episode. PPO failed to learn the behavior needed to exploit these sparse or delayed signals. In contrast, our policy search methods evaluate complete playouts and therefore do not face the same temporal credit-assignment problem. A PPO agent taking random actions is unlikely to encounter its first Dodgeball reward. Program search instead explores DSL programs in a language that explicitly represents the target and the ball, biasing it toward policies that interact with both.
\end{sloppypar}

\begin{table*}[t]
  \centering
  \scriptsize
  \setlength{\tabcolsep}{2.0pt}
  \renewcommand{\arraystretch}{1.08}
  \begin{tabular*}{\textwidth}{@{}l@{\extracolsep{\fill}}*{14}{r}@{}}
    \toprule
    Method
      & \multicolumn{1}{c}{ButtonCats}
      & \multicolumn{1}{c}{ButtonCheese}
      & \multicolumn{1}{c}{ButtonCheeseCats}
      & \multicolumn{1}{c}{Chase}
      & \multicolumn{1}{c}{Dodgeball}
      & \multicolumn{1}{c}{Football}
      & \multicolumn{1}{c}{Hide}
      & \multicolumn{1}{c}{Lawnmower}
      & \multicolumn{1}{c}{MouseCats}
      & \multicolumn{1}{c}{MouseCheeseCats}
      & \multicolumn{1}{c}{Pathfinding}
      & \multicolumn{1}{c}{ProtectTheChild}
      & \multicolumn{1}{c}{Seek}
      & \multicolumn{1}{c}{Vacuum} \\
    \midrule
    Search            & 21          & 12          & 7           & 6           & \textbf{12} & -1          & 1747          & 57          & 1088          & 11          & 5           & 20810          & \textbf{4} & 39 \\
    Coding agent      & 17          & 9           & 6           & 2           & 3           & -1          & 1675          & 52          & 758           & 6           & 6           & 16402          & \textbf{4} & 30 \\
    Agentic sketching & 24          & \textbf{14} & \textbf{12} & 8           & 9           & \textbf{0}  & \textbf{1858} & 60          & \textbf{1572} & \textbf{15} & \textbf{8}  & \textbf{22684} & \textbf{4} & \textbf{50} \\
    Sk. w/o antipatterns
                              & 22 & 14 & 9 & 8 & 9 & -2 & 1812 & 60 & 1266 & 11 & 6 & 20810 & 4 & 50 \\
    PPO               & \textbf{157}& 9           & 6           & \textbf{35} & 0           & \textbf{0}  & 581           & \textbf{66} & 595           & 14          & 2           & 11098          & 1           & 33 \\
    PPO avg. score    & 172.0 & 13.0 & 8.0 & 62.0 & 1.9 & 0.4 & 1481.4 & 90.6 & 3031.4 & 16.2 & 7.6 & 15277.5 & 5.4 & 61.2 \\
    \bottomrule
  \end{tabular*}
  \caption{Scores of the best policies found by each method. In our benchmark, a game's aggregate score is the minimum score across $N$ random seeds. For comparison, we also report the average PPO score across those seeds.}
  \label{tab:ppo-results}
\end{table*}

\begin{table*}[t]
  \centering
  \scriptsize
  \setlength{\tabcolsep}{2.0pt}
  \renewcommand{\arraystretch}{1.08}
  \begin{tabular*}{\textwidth}{@{}l@{\extracolsep{\fill}}*{14}{r}@{}}
    \toprule
    Method
      & \multicolumn{1}{c}{ButtonCats}
      & \multicolumn{1}{c}{ButtonCheese}
      & \multicolumn{1}{c}{ButtonCheeseCats}
      & \multicolumn{1}{c}{Chase}
      & \multicolumn{1}{c}{Dodgeball}
      & \multicolumn{1}{c}{Football}
      & \multicolumn{1}{c}{Hide}
      & \multicolumn{1}{c}{Lawnmower}
      & \multicolumn{1}{c}{MouseCats}
      & \multicolumn{1}{c}{MouseCheeseCats}
      & \multicolumn{1}{c}{Pathfinding}
      & \multicolumn{1}{c}{ProtectTheChild}
      & \multicolumn{1}{c}{Seek}
      & \multicolumn{1}{c}{Vacuum} \\
    \midrule
    Search            & 20 & 20 & 18 & 21 & 22 & 10 & 19 & 25 & 23 & 20 & 20 & 23 & 19 & 30 \\
    Sketching, subexpr & 17 & 18 & 14 & 18 & 20 & 9 & 15 & 20 & 17 & 17 & 20 & 36 & 16 & 22 \\
    Sketching, sketch  & 35 & 29 & 69 & 49 & 58 & 48 & 47 & 46 & 45 & 38 & 33 & 27 & 111 & 68 \\
    Coding agent       & 56 & 54 & 163 & 75 & 31 & 40 & 25 & 68 & 47 & 48 & 91 & 149 & 77 & 42 \\
    \bottomrule
  \end{tabular*}
  \caption{Maximum expression sizes used by Search, Agentic Sketching, and the Coding Agent. For Agentic Sketching, we report both the maximum size of the searched subexpressions and the maximum sketch size.}
  \label{tab:expression-sizes}
\end{table*}

\begin{table*}[t]
  \centering
  \scriptsize
  \setlength{\tabcolsep}{1.7pt}
  \renewcommand{\arraystretch}{1.08}
  \begin{tabular*}{\textwidth}{@{}l@{\extracolsep{\fill}}*{15}{r}@{}}
    \toprule
    Statistic
      & \multicolumn{1}{c}{\shortstack{Button\\Cats}}
      & \multicolumn{1}{c}{\shortstack{Button\\Cheese}}
      & \multicolumn{1}{c}{\shortstack{Button\\Cheese\\Cats}}
      & \multicolumn{1}{c}{Chase}
      & \multicolumn{1}{c}{Dodgeball}
      & \multicolumn{1}{c}{Football}
      & \multicolumn{1}{c}{Hide}
      & \multicolumn{1}{c}{Lawnmower}
      & \multicolumn{1}{c}{\shortstack{Mouse\\Cats}}
      & \multicolumn{1}{c}{\shortstack{Mouse\\Cheese\\Cats}}
      & \multicolumn{1}{c}{Pathfinding}
      & \multicolumn{1}{c}{\shortstack{Protect\\the Child}}
      & \multicolumn{1}{c}{Seek}
      & \multicolumn{1}{c}{Vacuum}
      & \multicolumn{1}{c}{Total} \\
    \midrule
    Search attempts       & 16 & 13 & 13 & 21 & 20 & 5 & 11 & 32 & 12 & 17 & 20 & 5 & 11 & 17 & 213 \\
    Valid searches        & 14 & 13 & 13 & 20 & 18 & 5 & 10 & 31 & 12 & 16 & 19 & 5 & 10 & 17 & 203 \\
    Direct policy evals    & 7  & 3  & 5  & 5  & 7  & 7 & 1  & 1  & 1  & 1  & 7  & 0 & 5  & 2  & 52  \\
    Distinct sketches     & 9  & 10 & 13 & 18 & 14 & 4 & 5  & 21 & 10 & 11 & 9  & 3 & 8  & 15 & 150 \\
    Bound increases (imm.) & 4 (3) & 3 (0) & 0 (0) & 2 (0) & 3 (2) & 0 (0) & 4 (4) & 9 (2) & 2 (0) & 4 (2) & 10 (6) & 1 (1) & 2 (1) & 2 (0) & 46 (21) \\
    \bottomrule
  \end{tabular*}
  \caption{Agentic sketching tool calls. Bound increases are revisits of the same sketch with a larger search bound (immediately consecutive revisits are shown in parentheses).}
  \label{tab:agentic-tool-calls}
\end{table*}

\subsection{Qualitative results}
In addition to the aggregate results above, our method discovers several policies that were unexpected to us. Beyond their role in our system, we think these policies are independently valuable to game developers. In the Pathfinding game, agentic sketching discovered the policy shown in Figure~\ref{fig:dsl_example}, which effectively navigates to a goal in a maze using only raycast distances, without graph search and without internal memory. In qualitative tests, we observed that this policy can escape narrow cul-de-sacs. This behavior appears to result from its maximization of raycast distance, which biases the agent toward directions with longer unobstructed paths. However, this local heuristic does not provide the global reasoning afforded by graph-based planning. Consequently, the policy may fail to escape wider cul-de-sacs, where locally favorable directions do not necessarily lead to an exit, and more complex zigzagging mazes, which require a coordinated sequence of turns that may temporarily move the agent away from the goal.
In the Chase game, both bottom-up search and agentic sketching found a policy that lets a slow robot reliably collide with a fast randomly moving target, again without planning.

%% file: 7-Conclusion.tex
\section{Conclusion}
We presented a benchmark and a synthesis method for learning programmatic policies in continuous games. The benchmark contains 14 games that share a single physical interface and expose behaviors such as pursuit, avoidance, navigation, occlusion, item use, and finite-turn-rate movement. To search this space, we introduced a compact DSL for reactive game policies, a bottom-up enumerator accelerated by antipatterns, and agentic sketching, which lets a coding agent propose high-level policy structure while symbolic search fills local holes exhaustively. Our experiments show that carefully designed symbolic search is a strong baseline, outperforming a pure coding agent while running on a laptop CPU, and that combining the two approaches produces the best results. The learned policies are compact, executable, and editable, although they are discovered from reward signals. These results suggest that programmatic policy synthesis can bridge the gap between hand-authored game AI and opaque learned controllers, offering game developers a practical way to generate behaviors that remain inspectable and controllable.

\subsection{Limitations}
The main limitation of our current system is the lack of numeric parameters in policies. Our DSL contains only one numeric constant, $2.5$, so the search cannot tune thresholds, radii, offsets, gains, or timing parameters for a particular game. This keeps enumeration small and makes policies easier to compare, but it also rules out many useful behaviors that depend on precise numeric choices. It is possible to combine symbolic enumeration with gradient-based optimization over numeric holes, but we did not find an efficient way to do so.

\subsection{Future work}
We currently hand-code antipatterns. A natural next step is to derive them automatically using static analysis or abstract interpretation, so the enumerator can prune redundant regions of the DSL without manual rules. Policy evaluation could also return richer information, such as rollout traces, failure states, state distributions, and intermediate game events, rather than only the final reward. For agentic sketching, the coding agent could create simplified versions of a game, analogous to curricula or unit tests, use them to discover useful policy fragments, and then transfer those fragments back to the original task.

%% file: dsl-grammar.tex
\begin{table*}[p]
\centering
  \caption{DSL grammar in a machine-readable form; game-declared sensors and templates are represented by typed names.}
  \label{tab:dsl-grammar}
  \begin{minipage}{0.96\textwidth}
  \small
\begin{verbatim}
program        ::= action

action         ::= (go vector)
                 | (goto object)
                 | use
                 | (if boolean action action)

boolean        ::= (< float float)
                 | (== object object)
                 | (aligned vector vector)
                 | (tile vector)
                 | game_expr<boolean>

float          ::= 2.5
                 | (x2 float)
                 | (* float float)
                 | (distance object object)
                 | (alignment vector vector)
                 | (gap vector)
                 | (raydist vector)
                 | (rundist vector template)

object         ::= self | none | game_expr<object>

template       ::= template_name

vector         ::= #0 | (dir object) | (left vector) | (right vector)
                 | (to object) | (from object) | (- vector vector)
                 | (maxDirection float)

game_expr<T>   ::= sensor_name<T>
\end{verbatim}
  \end{minipage}
\end{table*}